\documentclass[3p,authoryear,preprint,12pt]{elsarticle}
\makeatletter\if@twocolumn\PassOptionsToPackage{switch}{lineno}\else\fi\makeatother

\usepackage{graphicx} % Required for inserting images
\usepackage{tabulary,xcolor}
\usepackage{amsfonts,amsmath,amssymb}
\usepackage[T1]{fontenc}

\usepackage{ifluatex}
\ifluatex
\usepackage{fontspec}
\defaultfontfeatures{Ligatures=TeX}
\usepackage[]{unicode-math}
\unimathsetup{math-style=TeX}
\else 
\usepackage[utf8]{inputenc}
\fi 
\ifluatex\else\usepackage{stmaryrd}\fi

\usepackage{amsmath}
\usepackage{algorithm}
\usepackage{algorithmic}
\usepackage{multirow}
\usepackage{amssymb}
\usepackage{booktabs}
\usepackage{xcolor, soul}
\usepackage[hidelinks]{hyperref}
\usepackage{float}
\usepackage{microtype}
\numberwithin{equation}{section}

\usepackage{fancyhdr}
\begin{document}

\begin{frontmatter}

%% Title, authors and addresses

%% use the tnoteref command within \title for footnotes;
%% use the tnotetext command for theassociated footnote;
%% use the fnref command within \author or \address for footnotes;
%% use the fntext command for theassociated footnote;
%% use the corref command within \author for corresponding author footnotes;
%% use the cortext command for theassociated footnote;
%% use the ead command for the email address,
%% and the form \ead[url] for the home page:
%% \title{Title\tnoteref{label1}}
%% \tnotetext[label1]{}
%% \author{Name\corref{cor1}\fnref{label2}}
%% \ead{email address}
%% \ead[url]{home page}
%% \fntext[label2]{}
%% \cortext[cor1]{}
%% \affiliation{organization={},
%%             addressline={},
%%             city={},
%%             postcode={},
%%             state={},
%%             country={}}
%% \fntext[label3]{}

\title{Facial Chick Sexing: \\ An Automated Chick Sexing System From Chick Facial Image}

\author[inst1]{Marta Veganzones Rodriguez$^\dagger$}
\author[inst1]{Thinh Phan$^\dagger$ \footnote{$^\dagger$ indicates the same contribution}}
\author[inst2]{Arthur F. A. Fernandes}
\author[inst2]{Vivian Breen}
\author[inst2]{Jesus Arango}
\author[inst3]{Michael T. Kidd}
\author[inst1]{Ngan Le $^\ddagger$\footnote{ $^\ddagger$ indicates corresponding author}}

% \author[inst1]{Marta Veganzones Rodriguez \texorpdfstring{ $^\dagger$}{}}
% \author[inst1]{Thinh Phan \texorpdfstring{$^\dagger$}{} \footnote{\texorpdfstring{$^\dagger$}{} indicates the same contribution}}
% \author[inst2]{Arthur F. A. Fernandes}
% \author[inst2]{Vivian Breen}
% \author[inst2]{Jesus Arango}
% \author[inst3]{Michael T. Kidd}
% \author[inst1]{Ngan Le \texorpdfstring{$^\ddagger$}{} \footnote{ \texorpdfstring{$^\ddagger$}{} indicates corresponding author}}

\affiliation[inst1]{organization={Department of Computer Science and Computer Engineering},%Department and Organization
            addressline={1 University of Arkansas}, 
            city={Fayetteville},
            postcode={72701}, 
            state={Arkansas},
            country={USA}}

\affiliation[inst2]{organization={Cobb Vantress, Inc},%Department and Organization
            addressline={4703 US HWY 412 E}, 
            city={Siloam Springs},
            postcode={72761}, 
            state={Arkansas},
            country={USA}}
\affiliation[inst3]{organization={Department of Poultry Science},%Department and Organization
            addressline={1260 W. Maple, POSC O-114}, 
            city={Fayetteville},
            postcode={72701}, 
            state={Arkansas},
            country={USA}}

\begin{abstract}
%% Text of abstract
Chick sexing, the process of determining the gender of day-old chicks, is a critical task in the poultry industry due to the distinct roles that each gender plays in production.
While effective traditional methods achieve high accuracy, color, and wing feather sexing is exclusive to specific breeds, and vent sexing is invasive and requires trained experts. To address these challenges, we propose a novel approach inspired by facial gender classification techniques in humans: \textbf{facial chick sexing}. This new method does not require expert knowledge and aims to reduce training time while enhancing animal welfare by minimizing chick manipulation. We develop a comprehensive system for training and inference that includes data collection, facial and keypoint detection, facial alignment, and classification. We evaluate our model on two sets of images: Cropped Full Face and Cropped Middle Face, both of which maintain essential facial features of the chick for further analysis. Our experiment demonstrates the promising viability, with a final accuracy of 81.89\%, of this approach for future practices in chick sexing by making them more universally applicable.

\end{abstract}

%%Graphical abstract
\begin{graphicalabstract}
%\hl{Need a main figure}

\centering
\includegraphics[width=0.7\linewidth]{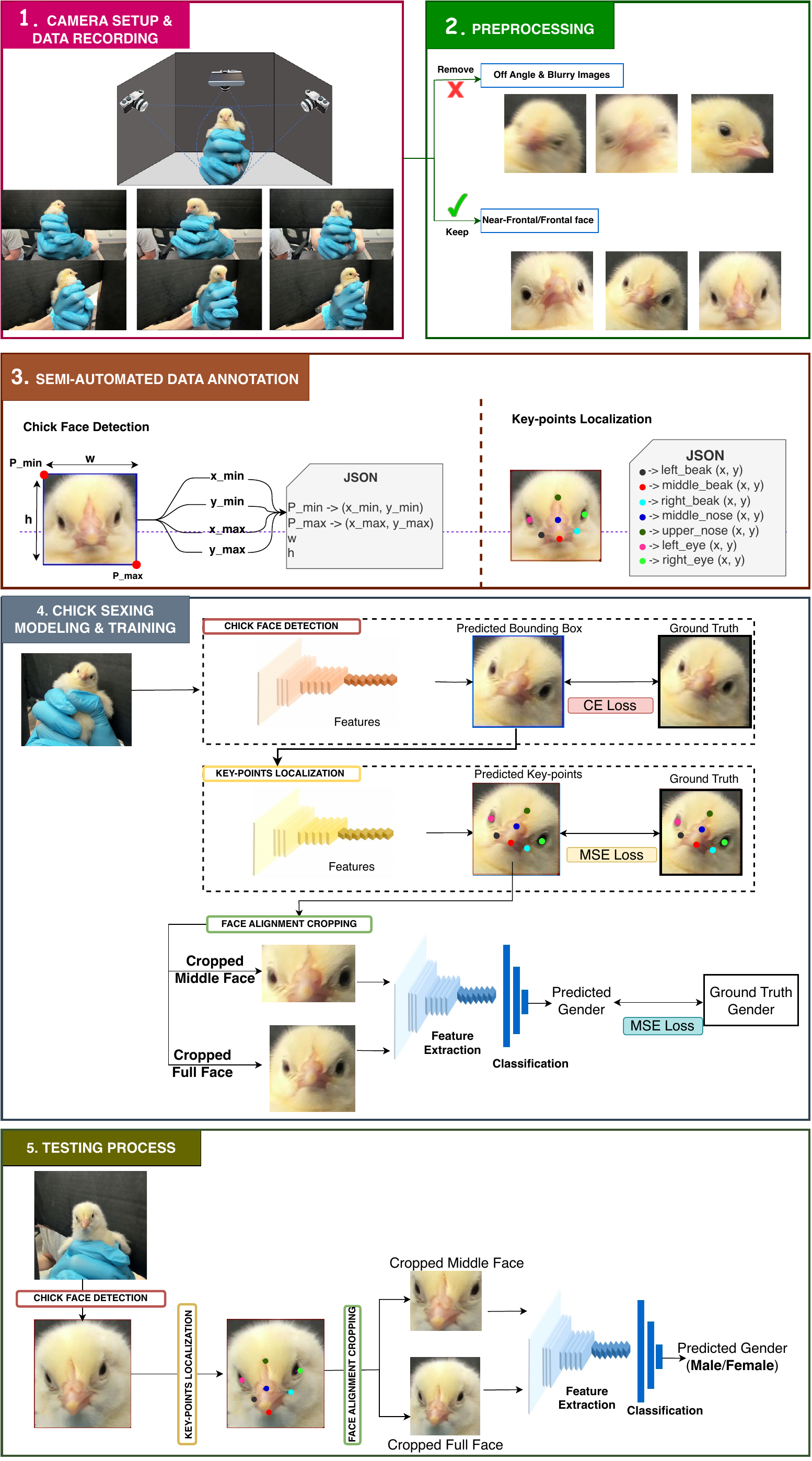}
\end{graphicalabstract}

%%Research highlights
\begin{highlights}
\item  \textbf{Dataset}: We obtained a dataset of facial images of one-day-old chicks, consisting of a total of 5,727 images captured from 184 female and 169 male chicks. The dataset has been carefully annotated by 2 poultry experts and 2 computer vision experts.
\item \textbf{Methodology}: We introduce an automated chick sexing system capable of accurately identifying the gender of day-old chicks. This system comprises of five components: chick face localization, facial chick keypoint detection, facial chick alignment, feature extraction, and gender classification. Designed as an end-to-end framework, our system provides a comprehensive solution for efficient and adequate chick sexing.
%\item \textbf{Pre-trained models and Code}: The pre-trained model, and source code of the proposed facial chick sexing are property of Cobb-Vantress and may be available for research purposes upon request. 
\end{highlights}

\begin{keyword}
%% keywords here, in the form: keyword \sep keyword
Chick Sexing, Poultry \sep Non-invasive \sep Computer Vision (CV) \sep Gender Classification \sep Deep Learning (DL)
 % , Transformer \sep
%% PACS codes here, in the form: \PACS code \sep code
% \PACS 0000 \sep 1111
% %% MSC codes here, in the form: \MSC code \sep code
% %% or \MSC[2008] code \sep code (2000 is the default)
% \MSC 0000 \sep 1111
\end{keyword}

\end{frontmatter}

\section{Introduction}
% no \PARstart
%\hl{all citations has problem! }

The classification of male or female broilers, namely chick sexing, is the crucial initial step for raising sexed flocks of broilers, female-only layers, and other practices to increase efficiency and profitability in the poultry industry. For mass broiler production, male broilers are preferred due to their higher meat yield, growth rate, food intake, and uniformity in size \cite{weimerskirch2000sex, alin2019non}. Regarding higher meat quality, there is no evidence that one gender is a better choice than the other \cite{abdullah2010growth}. In the egg production industry, if the chicks are not gender-classified in the early stages and raised until visible cues are observable, there would be the burden of the feeding the undesired males \cite{he2019simple}. Therefore, several traditional methods have been devised and applied to day-old chicks to avoid this. Wing feather sexing technique is based on the growth rate and length of the primary and secondary wing feathers \cite{dakpogan2012effectiveness}. Unfortunately, this can only be applied to specific chicken breeds due to its genetics \cite{jia2023review}. Vent sexing is a comprehensive technique introduced by the Japanese in the 1930s, involving examining the cloaca area to detect gender-exclusive features \cite{biederman1987sexing}. Despite the proven high accuracy, vent sexing requires a long training period and expert knowledge to recognize subtle decisive clues between the genders. Moreover, it can hurt the chick if not carried out properly. Considering those challenges, there is a need for a new non-invasive method in this field. This need for innovation has led to the exploration and implementation of machine learning and computer vision techniques offering a transformation of the industry.
%. %\hl{in this paragraph, you can add more detailed analysis on vent chick sexing and feather chick sexing, particularly focus on the process and their drawbacks.}

%The next one: need to describe a little bit human face gender classification. 
Regarding humans, many researchers have proposed to classify gender through body appearance, voice, gait, and facial features, from these, facial features have been the most prevalent. Facial features are one of the most reliable characteristics for both human beings and computer systems to carry out the task of gender recognition. In earlier approaches, the process of facial gender classification could be summarized into four main steps: face detection, image pre-processing and enhancement, facial feature extraction and classification. The last two steps played important roles in the performance of the method. Principal Component Analysis (PCA) \cite{buchala2005role}, Gabor wavelet \cite{leng2008improving}, Scale Invariant Feature Transform (SIFT) \cite{zhang2011hierarchical}, Local Binary Patterns (LBP) \cite{li2012gender} and Histograms of Oriented Gradients (HOG) \cite{lee2013novel} were some of the popular feature extractor to obtain facial features. They were then fed to machine-learning classifiers such as Random Forest \cite{khan2019automatic}, Support Vector Machine (SVM) \cite{lee2010automatic} or Adaboost \cite{afifi2019afif4} for the final prediction. Thanks to the rapid development of deep learning methods, Convolutional Neural Network (CNN) not only reduced the workload by merging the two mentioned tasks but also boosted the accuracy by considerable margin \cite{sumi2021human}. With the current large quantity of training data and the advancement of deep learning models, human facial gender classification has already not been a significant issue in recent years. 

%\hl{Emphasize on the advancements of facial chick sexing }

However, whether the human face detection and gender classification technique could be transferred to young chickens is still an open question. The human face and the chicken face do not share similar structures. While adult chickens can be discerned by the comb, day-old chicks seemingly look alike without noticeable visual clues. Motivated by this issue, our work aims to explore and evaluate the adaptation of the human-based method on the faces of day-old chicks and evaluate if this is a viable approach for chick gender identification. Building on this, the application of facial chick sexing introduces several advancements. Contrasting with traditional methods, this new approach is non-invasive, does not require expert knowledge, and reduces stress on the chicks, thereby enhancing animal welfare.

%\hl{The contributions of our proposed methods are:}
Our contributions are summarized as follows:

\begin{itemize}
    % % \item Development of a multi-view recording system: we designed a system involving three cameras in three angles within a black box to reduce illuminations, enabling consistent image capturing.
    % \item Acquire the frontal/near-frontal chick face face dataset: our methodology involves selecting and extracting frames with clear frontal/near-frontal views of chick faces, creating a focused dataset.
    % \item Develop a computational model for chick sexing: we propose a model that adapts the four steps used in human facial gender classification to chicks. This includes face detection, landmark detection, facial alignment and concludes with the chick gender classification. 
    \item \textbf{Dataset}: A dataset containing a total of 5,727 images of one-day-old face chicks from 184 female and 169 male chicks. The dataset was carefully annotated by two computer vision experts and two poultry experts. Both chick face location, facial keypoints, and gender were annotated.
    \item \textbf{Methodology}: We introduce an automated chick sexing system capable of accurately identifying the gender of day-old chicks. This system comprises five components: chick face localization, facial chick keypoint detection, facial chick alignment, feature extraction, and gender classification. Designed as an end-to-end framework, our system provides a comprehensive solution for efficient and precise chick sexing.
    % \item \textbf{Pre-trained models and Code}: The pre-trained model, and source code of the proposed facial chick sexing are property of Cobb-Vantress and may be available for research purposes upon request.

\end{itemize}

To the best of our knowledge, this work represents the pioneering effort to explore the automated recognition of chick sexing through computer vision technologies, leveraging facial analysis to discern the gender of day-old chicks based on their facial features.

\section{Related work}

This section reviews the progress from conventional sexing techniques, including vent and wing feather sexing, to the application of deep learning models for the automation of the existing manual works. It then transitions to a review of the human gender recognition methods. The discussion concludes with our proposed method inspired by these advancements in human gender recognition, tailored to develop a new approach for chick sexing.

\subsection{Chick Sexing}
%Listing all related work and then mention some drawbacks of those approaches
\subsubsection{Feather Sexing}
Wing feather sexing is an easier and simplified method to classify day-old chicks by gender. This method consists of analyzing the length of the primary and secondary wing feathers. If they have a gene for fast feathering, meaning that the primary feather is longer and thicker than the covert feather, it is determined to be female. On the contrary, if they have a gene for slow feathering, meaning that the primary and covert feather have the same length meaning, it is determined to be male \cite{england2023influence}. The K gene, located in the Z chromosome, determines the growth feather rate in chicks. Some of the advantages of implementing this method are that it does not require trained experts to reproduce this classification, it is less invasive for the chick, it is also safer and reasonably fast. On average, a trained person can sex between 2,000-2,500 chicks per hour \cite{nandi2003sex}. This process acquires between 80\%-95\% accuracy depending on the type of breeds. However, it cannot be used for all type of breeds and accuracy can vary. 

Initial investigations, such as \cite{jones1991edge}, explored machine vision techniques for feather sexing, using edge detection algorithms to automate the gender identification process. While these initial methods showed promise, they achieved lower accuracies ranging from 50\% to 89\%, with the highest accuracies achieved through the application of discriminant analysis. Recent advances in machine learning, specifically CNNs, have revolutionized the approach to chick sexing. Automated feather sexing using CNNs has proven effective, yielding accuracy rates exceeding 80\%, significantly higher than the 50\% achieved by their experts \cite{soliman2023day}. These approaches apply image processing techniques to datasets of chick images, where networks analyze and recognize patterns to accurately determine the gender of unseen chicks. This automation significantly reduces human error and fatigue, enhancing efficiency, reliability and consistency on the results.

% feather sexing chicks per hour reference (Organisational framework for hatcheries): Avian sex determination and sex diagnosis. (2003). World’s Poultry Science Journal, 59(1), 5–64. https://doi.org/10.1079/WPS20030002
%https://www.tandfonline.com/doi/pdf/10.1079/WPS20034 check this

\subsubsection{Vent Sexing}
The technique widely used in large-scale hatcheries for gender determination is vent sexing. This technique compared to wing feather can be applied to any type of chicks. It consists of turning the chick up side down and gently exposing the cloacal area. Before the exposure, it is necessary to expel any fecal debris from the vent that may interfere with the observation \cite{kalejaiye2021poultry}. To identify the sex of the chick, we focus on the lower cloacal area. If we find a white small genital bump, it will be determined as male. Otherwise, it will be determined as female. Improper exposure of the chick's genital organs can cause minor or major injuries, or transmission of viruses causing early death \cite{phelps2003automated, iswati2020sexing, england2021sexing}. In addition, the complexity of this technique requires a high level of precision to detect the unique characteristics of each gender. Therefore, this complex technique needs to be performed by expert trainers, requiring approximately 18 months of training. On average, a trainer can sex between 800-1,000 chicks per hour \cite{nandi2003sex}. Despite the drawbacks, it presents one of the better accuracies among other techniques, achieving a high accuracy of 97\% \cite{biederman1987sexing}.

%In 2023, a revolutionary chick sexing identification machine was released to the world being the first commercial machine doing automated vent sexing \cite{16}. This machine employs Artificial Intelligence algorithms to determine the gender of the chicks. It can be used to identify the gender of white, coloured broilers, layers, and turkey. The process consist on an operator positioning the chick in the detection region, where no contact sensors automatically capture images for identification reducing the time problem into less than 3 seconds and obtaining 98.5\% accuracy. Moreover, the machine only requires a five-day training period, no need for experts and it can process 1,000 chicks per hour. This not only enhances efficiency but also significantly reduces the labor required, addressing some of the major challenges of manual sexing methods. This innovative machine could potentially redefine standard operational practices and improve both productivity and animal welfare in the industry.

In 2023, \cite{xiashu2024automatic} revolutionized the field of chick sexing identification by developing and producing the world's first commercial chick sexing machine. This innovative machine leverages AI algorithms that uses millions of images from almost 30 different species to determine the gender of the chicks, including white-feathered broilers, layers, yellow feathered broilers, and variegated broilers. With only 5 days of training, operators learn to expose the vent towards the detection region of the camera, leaving the machine to automatically scan and capture images for automatic gender identification, reducing the entire process into 3 seconds per chick with an accuracy of 98.5\%. Moreover, it handles 1,000 chicks per hour and help producers to reduce the intense training labors and costs and minimizing the stress on the chicks as the process is shorter. However, it remains an invasive method as it requires exposing the vent.

The integration of state-of-the-art CNN methods into the chick sexing process represents a significant advance in this field, setting the stage for more accurate, less need for experts and less invasive classification techniques in the future.

\subsection{Human Facial Gender Classification}
%\hl{Please leverage 3.1 in this work }
%https://link.springer.com/article/10.1007/s11042-022-12678-6

%\hl{At the end, you can clarify your work. "In this work, we leverage human gender recognition and develop a chick sexing human chick face.}
%Recent studies have reported different perspectives by implementing machine learning and computer vision methods in this field, offering potential solutions to time-efficiency challenges.  

Human facial gender recognition has been a significant topic in the research world for years. This interest has been driven by the advancements in deep learning techniques and the increasing availability of large-scale datasets. Initial efforts were based on traditional machine learning methods, utilizing feature extraction techniques such as PCA \cite{buchala2005role}, Gabor wavelets \cite{leng2008improving} and LBP \cite{li2012gender} combined with SVM \cite{lee2010automatic}, Random Forest \cite{khan2019automatic} and Adaboost \cite{afifi2019afif4} classifiers.

Recent advancements have seen a shift from traditional machine learning methods towards deep learning models \cite{gupta2023single}, particularly Convolutional Neural Networks (CNNs). Initially, \cite{antipov2016minimalistic} utilized simplified CNN architectures, modifying their depth and showing that shallower CNNs still achieve high accuracy on large dataset without relevant performance loss. Furthermore, \cite{jia2016gender} demonstrated that deeper CNNs outperform methods like SVMs with LBP features when trained on large and noisy datasets. They also include more contextual information about the face region improving accuracy, achieving 98.90\% on the LFW dataset. Meanwhile, other researchers \cite{simanjuntak2020fusion} combined shifted filter responses (COSFIRE) with CNNs to reduce the error rates by more than 50\%. Additionally, \cite{d2019gender} effectively handled gender classification when variations in lighting, pose, ethnicity, and expressions were applied, utilizing VGG-Face Deep Convolutional Neural Network (DCNN) for feature extraction.

Different from the existing chick sexing methods, we propose a new technique that focuses on the face of the chicks, leveraging human gender recognition processes. This novel approach aims to explore a non-invasive and semi-automated solution for the poultry industry. 

\begin{figure}[!h]
    \centering
    \includegraphics[width=0.8\linewidth]{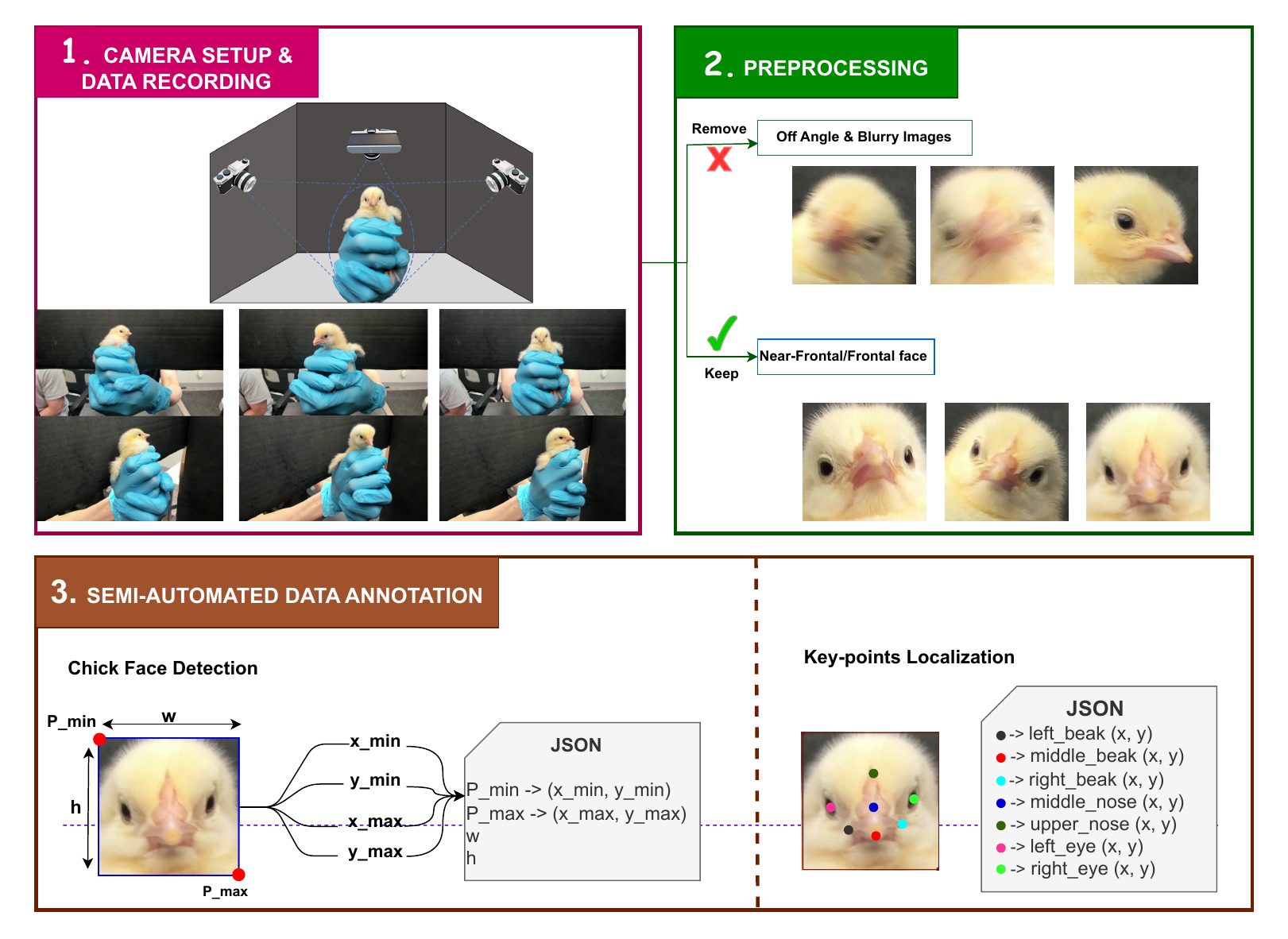}
    \caption{\textbf{Data acquisition and annotation process}. Our acquisition process involves setting up three cameras for data recording and pre-processing to eliminate low-quality frames and videos. The annotation process is semi-automated, starting with manual annotations to create an initial dataset, followed by model training, prediction and iterative corrections to efficiently annotate the entire dataset.
%Our acquisition process contains two steps of camera setup, data recording and preprocessing to elmimate all the low quality frames/videos. Our nnotation process is setup under a sermi-automated manner, ... 
    }
    \label{fig:Data_acquisition}
\end{figure}

%\hl{Different from the existing work, we ..... }
%Different from the existing work, we propose a new chick sexing technique applied to the face part of the chicks, leveraging the human gender recognition process.
%The poultry industry has the potential to benefit greatly from recognition technologies, as demonstrated by the proficient use of computer vision in the identification of chicks. This brings up a parallelism with facial recognition for humans, which has advanced significantly in the last several years. The human facial recognition \cite{17, 18} process involves the detection of faces in images or videos. Once the face is detected, it applies some algorithms such as the distance between the eyes, the shape of the jawline, and other unique facial landmarks, comparing them with a database of human faces. This pilot approach of a combination between face detection and alignment could potentially be adapted for use in animal facial recognition, including the poultry industry.

\section{Data Acquisition \& Labeling}
%\hl{All steps related to Cameras setup, Data collection}

\subsection{Overview}
As illustrated in Figure \ref{fig:Data_acquisition}, we build a recording system that captures the chick from three distinct angles. Because we have three views within a frame, we only need to select the frames with one high-quality frontal or near-frontal view of the chick's face. Following methods used in human facial gender classification, we require only the chick's face for further analysis. Consequently, given the variability in the chick's head position, we determine several keypoints to align the face straightforwardly.   

\subsection{ Multi-view Data Collection}

The data for this project was supported by one of the Cobb-Vantress facilities, where we were provided with 400 pre-sexed chicks, of which 200 were males and 200 were females. Since the chick's gaze is not always directed toward the camera and it is difficult to adjust its head position to fully face the front camera view, we developed %come up with 
a video recording system to increase the likelihood of capturing the chick's frontal facial features. As shown in Figure \ref{fig:Data_acquisition}, our system consists of three Logitech c920 cameras placed at the center and two corners. To minimize the negative impact of the background during the recording process, the cameras are set up inside a box with sides covered in black and LED lights on the roof. This design reduces unwanted illumination reflections, shadows, and distracted objects. Each video is approximately 10 seconds in length, recorded at 30 frames per second in 2K resolution. The three camera views are stacked vertically in every frame. During recording, we rotated the chick to capture more frontal or near-frontal views of its head. %The difficulties encountered in capturing the optimal frontal frame, along with the selected frames demonstrating a clear frontal view, are illustrated in Figure \ref{fig:quality_frame_selection}.

%\begin{figure}[h]
%    \centering
%    \includegraphics[width=0.5\textwidth]{Figs/Multiview.png}
%    \caption{Multi-view camera setup that illustrates the difficulty to capture frontal frames, and the selection of the frames with a clear frontal face. \hl{you can comebine this with Fig. 3}}
%    \label{fig:multiview_frames}
%\end{figure}

\subsection{Quality Frame Selection}
\begin{figure}[h]
    \centering
    \includegraphics[width=1.\textwidth]{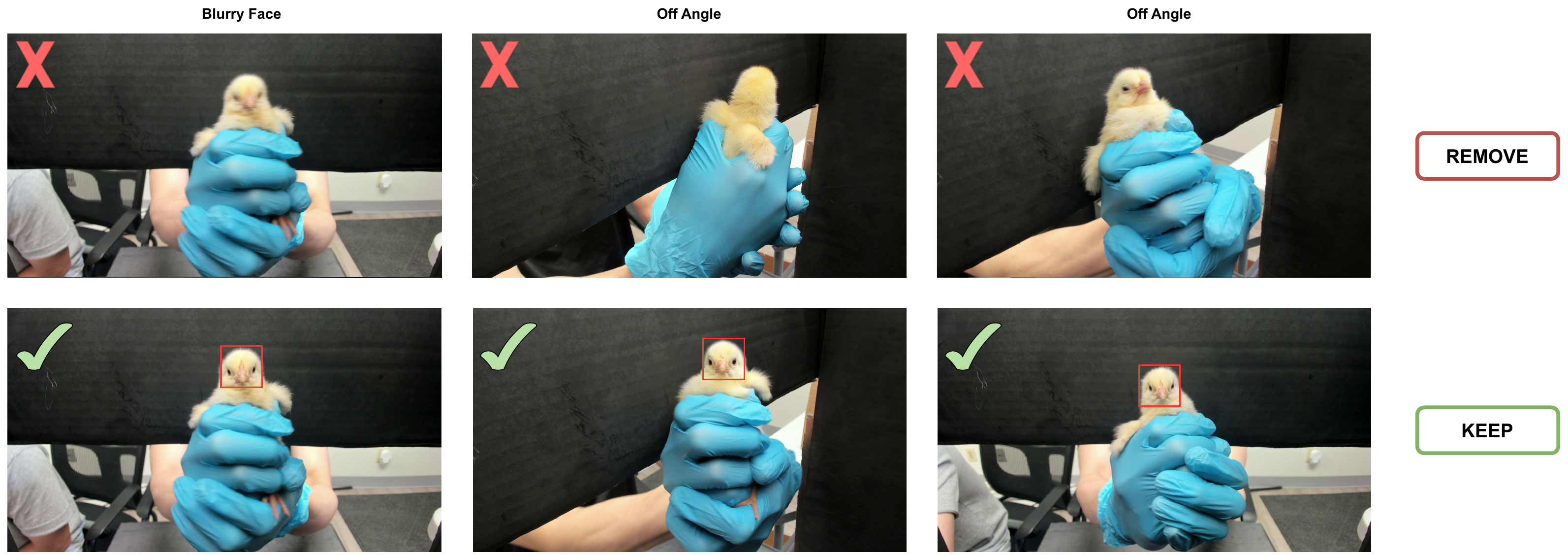}
    \caption{Comparison between the two frames on the left that represent an example of frames that do not meet the standards and the one on the right that represents of a frame that satisfies the standards}
    \label{fig:quality_frame_selection}
\end{figure}
Since we only need the frontal or near-frontal views of the chick's head, only the view per frame that satisfies the requirement is selected. Additionally, even if the view is frontal or near-frontal, we skip the blurry or off-angle images, retaining only those that are clear and of high quality, as depicted in Figure \ref{fig:quality_frame_selection}.

Due to these criteria, 47 chick videos that did not meet the standards were discarded from the study, of which 31 were males and 16 were females. After filtering the initial dataset, a total of 5,727 sequences of images were selected, comprising 2,440 images of 169 male chicks and 3,287 images of 184 female chicks, which constitute the test set for further analysis. The data is structured as outlined in Table \ref{table1}.
%\hl{this is the test set}

\begin{table}[h]
\centering
\caption{Summary of the distribution of chick IDs and images by gender, including ranges of images per ID.}
\begin{tabular}{l|c|c|c}
\toprule
       & \# chick IDs    & \# images  & \#images per chick ID \\ \midrule
Female &    184    &   3,287    &      1-89          \\ \midrule
Male   &    169    &   2,440    &      1-135                           \\ \midrule
\textbf{Total}  &    \textbf{353}    &   \textbf{5,727 }   &      \textbf{1-135 }      \\ \bottomrule
\end{tabular}
\label{table1}
\end{table}

\subsection{ Semi-Automatic Data Annotation}

%Considering there is no prior existing dataset to train our models with, it is necessary to do manual annotations.\hl{Inspired by techniques used in human gender classification, our goal is to ensure all chick faces to be positioned straightforwardly}. With the help of the image annotation and labeling tool, LabelMe \cite{labelme2016}, we draw a bounding box around each chick's face, discarding the background for further analysis. Additionally, we also annotate seven facial key-points located as follows: 1. upper nose, 2. middle nose, 3. right eye, 4. right beak, 5. middle beak, 6. left beak and 7. left eye. These key-points are crucial for achieving our goal of aligning the faces, thereby enhancing the training outcomes. To reduce this time consuming labor task and consequently, accelerate this process with the remainder of the images, we developed a semi-automatic data annotation process supported by two Deep Learning models. These models are initially trained on a train set of 2,315 manually annotated images to predict the outputs of the unseen images. This train set is not separated by gender or chick ID, as the primary objective is to enable the models to predict the location of the face and key-points, without the need for gender or ID distinction.
%The images compromising the test set are structured as outlined in Table \ref{table1}.

As mentioned in section 3.1, we only need the aligned and cropped chick faces to train the gender classification model. However, the original images consistently include redundant background, and the chick's face is not always positioned straightforwardly within the frame. Currently, there is no existing tool to correctly detect and align the chick's face. Therefore, we initially built a small dataset with bounding boxes around the chick faces and crucial keypoints on their faces using the image annotation tool LabelMe \cite{labelme2016}. The bounding box locates the chick's face in the image and the keypoints are used for further chick head alignment. These keypoints include the upper nose, middle nose, right eye, right beak, middle beak, left beak and left eye, as shown in Figure \ref{fig:Data_acquisition}. Given the large amount of data, as detailed in Table \ref{table1}, we opted for a semi-automatic data annotation procedure. The annotation process is a cycle, in which we initially manually annotate a smaller part of the data (2,315 images) without distinguishing by gender or chick ID. Subsequently, we feed this data into the model for training and allow the model to make predictions on unlabeled images. The predicted information is then revised and corrected, making them the ground truth data. The model is updated with both the old and revised data and used to predict on new data again. This process is repeated until the entire dataset is annotated. By using this technique, we reduce the time and manual effort required for annotation. %The data is structured as outlined in Table \ref{table1}
% less time-consuming

%\hl{As mentioned in section 3.1, we only need the aligned and cropped chicken faces to train the gender classification model. However, the original image is mostly occupied by unwanted background and the chicken face's head positions are not always straightforward. There is no existing tool to detect the chicken face and align it correctly. Therefore, in the beginning, we build a small dataset with bounding boxes around chicken faces and crucial keypoints on their faces. The bounding box locates the chiken face in the image and the keypoints are used for further chicken head alignment. We have large amount of data, so we opt for semi-automatic data annotation. The annotation process is a cycle, of which in the beginning we manually annotate small part of data. Subsequently, we feed this data to model for training and let the model predict on unlabeled images. The predicted information is revised and corrected, making them the groundtruth data. The model is updated with old and revised data and predict on new data again. This process is repeated until the whole dataset is annotated. By using this technique, the annotation is less time-consuming and ...   }

%\hl{At the end of this section, you can provide a table, which contains all data including \# chicks regarding each gender, \# images, \# images/ID. We partition the data into train set and test set under the condition that IDs in the test set is not in the train set.}

\begin{figure*}[h]
    \centering
    \includegraphics[width=0.8\textwidth]{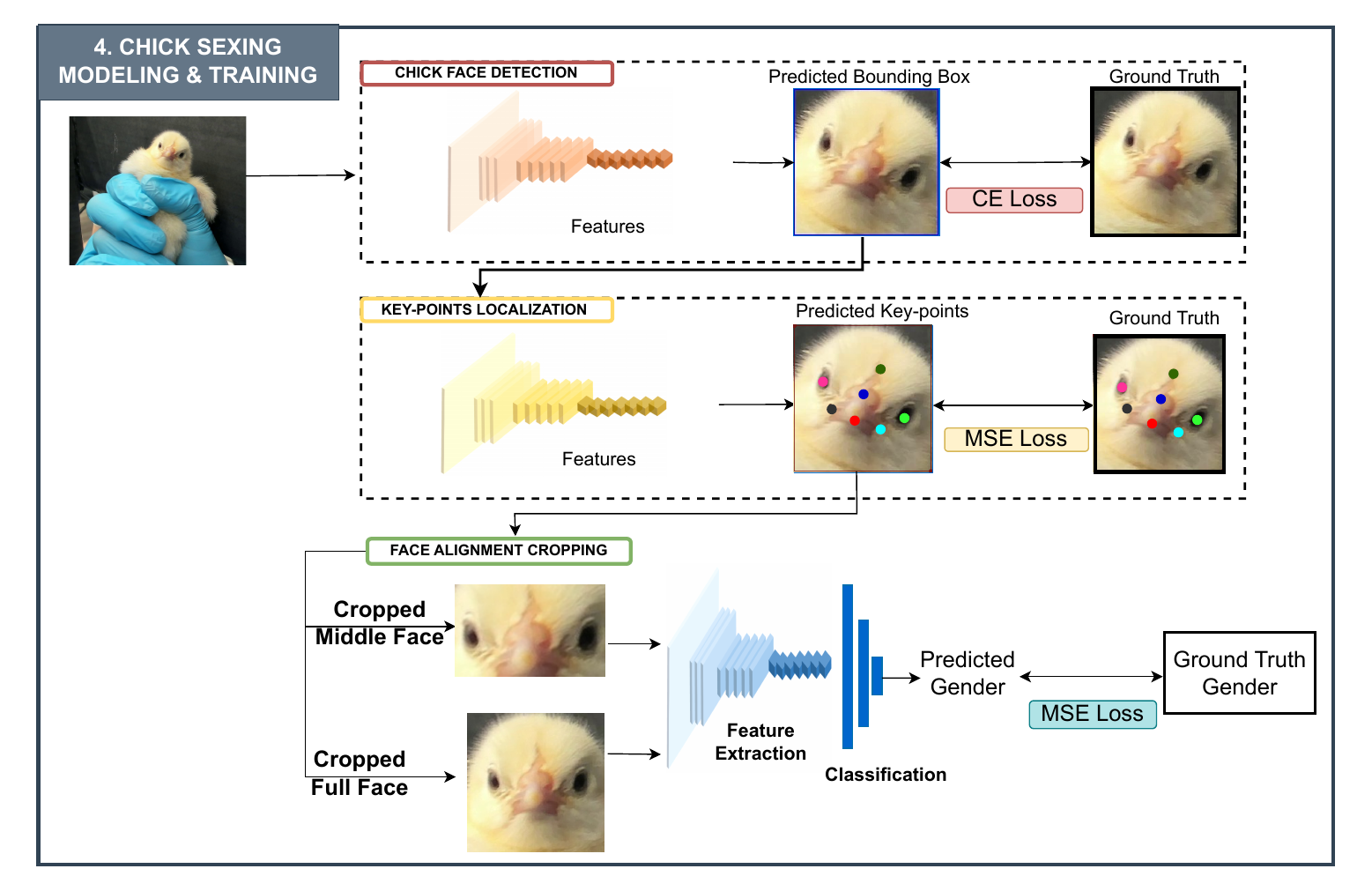}
    \caption{Schematic of the gender classification procedure for day-old chicks}
    \label{fig:schematic-gender-classification-procedure}
\end{figure*}

\section{Chick Face Recognition}
Following the four stages of the human facial gender recognition process, we leverage this framework to our facial chick sexing procedure, as shown in Figure \ref{fig:schematic-gender-classification-procedure}:
%Following this alignment, the process concludes with testing two cropping methods on the chick's face for the facial gender classification.

%\subsection{Bounding Box Detection Model} 
\subsection{Chick Face Localization}

To accurately localize and create the bounding boxes around the chick's face, we utilized the pre-trained open-source You Only Look Once (YOLOv5) \cite{jocher2022ultralytics} object detection model. This model, originally trained on the COCO dataset \cite{lin2014microsoft}, was chosen because of its high object detection accuracy. It was further trained on manually annotated bounding boxes and performed inference on a new set of images. The model was configured with an input image size of 640 pixels, a high confidence threshold of 0.8 for object detection, and an intersection over union (IoU) threshold of 0.5 to be considered a true positive detection. Non-maximum suppression was applied as the last step, refining multiple bounding box predictions to the best one, ensuring one detection per object. The application of this model resulted in a performance of 99\% accuracy in bounding box predictions on unseen images. There were only a few exceptions in these predictions requiring minimal vertical or horizontal fine-tuning, demonstrating the effectiveness of the selected model.

\subsection{Keypoints Detection}

%\hl{HRNet is for object detection, please provide more details on how to utilize it to detect keypoints}

Based on the resulting cropped chick face according to the predicted bounding box, we adapted and trained the High-Resolution Network (HRNet) \cite{wang2020deep} model to predict facial landmarks. These facial landmarks were determined using the manual key-points annotations from LabelMe \cite{labelme2016}. Although HRNet is widely known for object detection and human pose estimation, we utilized this network for our facial keypoint detection, as human pose estimation similarly requires the localization of human body joint keypoints such as shoulders and knees. Additionally, the preservation of high-resolution representations of the image throughout the entire network allows for precise keypoint detection. This model was originally trained on the large-scale COCO human pose detection \cite{lin2014microsoft} and keypoint detection train2017 benchmark \cite{wang2020deep}. %The main reason of utilizing this HRNet model architecture is that it maintains the high resolution representations of the image throughout the entire network. 
We utilized a pre-trained model on human data and then trained the model specifically for the chick face framework to enhance the accuracy of our predictions. To achieve this, we customized the weights and modified the final layer of the HRNet model to output the seven specific facial keypoints desired from the input image. %Following this approach, we customized the weights of the model specifically for these features to improve accuracy. \hl{This procedure allows us to test the model efficacy on a set of unseen images during the training phase.} 
The application of this model resulted in highly accurate keypoint predictions on the eyes, upper nose, and middle beak. However, after supervision, we observed less accurate results on the left and right sides of the beak, which require further manual fine-tuning in those keypoints. Despite the lower localization accuracy on some keypoints, the model's predictions significantly reduced the time needed for annotating these keypoints compared to manual annotations.

%From the manually annotated key-points on LabelMe \cite{19}, we adapted and trained the High-Resolution Network (HRNet) \cite{21} pose estimation model, originally trained over the benchmark COCO train2017 \cite{22} key-point detection and segmentation dataset for human pose estimation, to predict facial key-point features from chicks. Inspired by the model's architecture, which maintains the high resolution representations of the image throughout the entire network, we tailored our approach by employing a custom-trained HRNet model with weights specifically adjusted for chick key-point detection and applied bounding box normalization to extract scale and center information for image processing. This procedure allowed us to test the model efficacy on a set of unseen images during the training phase. The application of this model resulted in highly accurate key-point predictions on the eyes, upper nose, and beak but produced less accurate result for the left and right sides of the beak, indicating a need for additional time for fine-tuning those key-point results. 

\subsection{Face Alignment} 
%\hl{Add more calculations}
%Add that this is calculated manually 
To adapt the methodology from human facial analysis to day-old chicks, we developed a face alignment algorithm based on manual calculations of the geometry of specific landmarks. Initially, we calculate the midpoint between the left and right eye keypoints using the midpoint formula.
%By calculating the midpoint between the left and right eye key-points and using this midpoint as a pivot for rotation, we ensure that all chicks are presented in a uniformly frontal position prior classification. The midpoint between the eyes is calculated using the Midpoint formula.
\begin{equation}
M_1(x_3, y_3) = \left( \frac{x_1 + x_2}{2}, \frac{y_1 + y_2}{2} \right)
\end{equation}

However, since the midpoint is calculated in the world image coordinate system, which represents positions in the image relative to real-world dimensions and geometry rather than just pixel positions, we further adjust it relative to the bounding box of the chick's face. This serves as the pivot point for subsequent transformations, ensuring that rotation and scaling are centered on the chick's face. The adjusted midpoint is calculated as follows, where \textit{X} and \textit{Y} represents the top left coordinates of the bounding box. 
\begin{equation}
M_2(x_a, y_a) = \left( M_1 - X, M_1 - Y \right)
\end{equation}
This geometric calculation is critical for determining the angle of rotation $\theta$, derived from the vertical $(\Delta y)$ and horizontal $(\Delta x)$ differences between the eyes, to align the chick's face frontally, where $\Delta y = y_{2} - y_{1}$ and $\Delta x = x_{2} - x_{1}$. 
\begin{eqnarray}
 {\Theta} & = &  degrees(arctan2(\Delta y, \Delta x))
\end{eqnarray}
Using this calculated angle, we construct a transformation matrix, which allows us to rotate and scale the image around the adjusted midpoint, where $\alpha = scale * cos(\theta)$ and $\beta = scale * sin(\theta)$, with 1 as the scaling factor. 
\begin{equation}
\left[
\begin{array}{ccc}
\alpha & -\beta & (1-\alpha)\cdot M_2[0] - \beta\cdot M_2[1] \\
\beta & \alpha & \beta\cdot M_2[0] + (1-\alpha)\cdot M_2[1] 
\end{array} 
\right] 
\end{equation} 
Post-rotation, we transform the bounding box coordinates to fit the newly aligned face, overlaying the transformed bounding box onto the aligned image. Subsequently, we adjust the bounding box containing the chick's face to minimize the content of the natural background within the bounding box area. This process ensures that all chicks are presented in a uniformly frontal or near-frontal position prior to classification.

\begin{figure}[h]
    \centering
    \includegraphics[width=1.\textwidth]{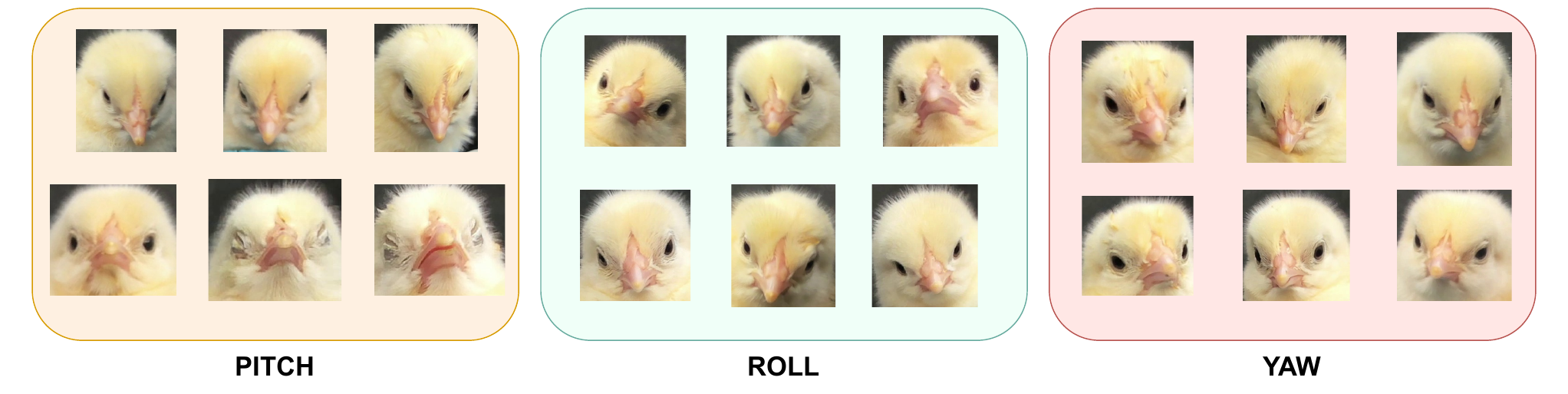}
    \caption{Chick Facial Pose Estimations: Cases of Roll, Pitch, and Some Instances of Yaw} 
    \label{fig:chick_poses}
\end{figure}

In addition to geometric alignment, we consider facial pose estimation, focusing particularly on the three cases of picth, yaw, and roll. However, we primarily consider roll and pitch cases, and some cases of yaw, as shown in Figure \ref{fig:chick_poses}. During the manual annotation process, we exclude yaw cases where the right and left beak, and right and left eye features are not visible within the frame to the camera. This ensures that the alignment process is robust and that the facial features are optimally positioned for further analysis.

%Through geometric analysis of the landmarks, we developed a face alignment algorithm to ensure all chicks appear in the most frontal position before being input into the computational classifier. This is achieved by calculating the midpoint between the left and right eye key-points and using it as the rotation pivot point. Subsequently, we refined the initially detected bounding box by trimming the chick's face to the maximum extent minimizing the amount of greyish background caused by light reflections in the black box.  

%\hl{This should be move to the Feature Extraction}

\subsection{Data Preparation}
Two different sets of input images of the chick's face are investigated and examined in this work: Cropped Full Face images and Cropped Middle Face images. These types were chosen to understand the effect of different facial regions on our analysis.
%Two different chick face feature extraction approaches are investigated and examined in this works: cropped full face feature and cropped middle face feature.
%going to provide two ways to extract the chick face feature by: 

%Prior to the final classification step, we consider two sets of images to be input into the model for further evaluation: Cropped Full Face and Cropped Middle Face.

{\textbf{i. Cropped Full Face}}

The first set of imagery inputs centers around the entire chick face area within the bounding box. This set consists of the resulting images obtained after the facial alignment procedure. 

%The first set of imagery inputs includes images exactly as they are after the face alignment, specially focusing on the full face part area. The adjustment of key-points conforms to the final aligned image. 

%\hl{This is just a input data, NOT feature}

{\textbf{ii. Cropped Middle Face}}
%The second chick face feature extraction aims to specifically focus on the salient chick's seven main features as outlined. 

\begin{figure}[h]
    \centering
    \includegraphics[width=1.\textwidth]{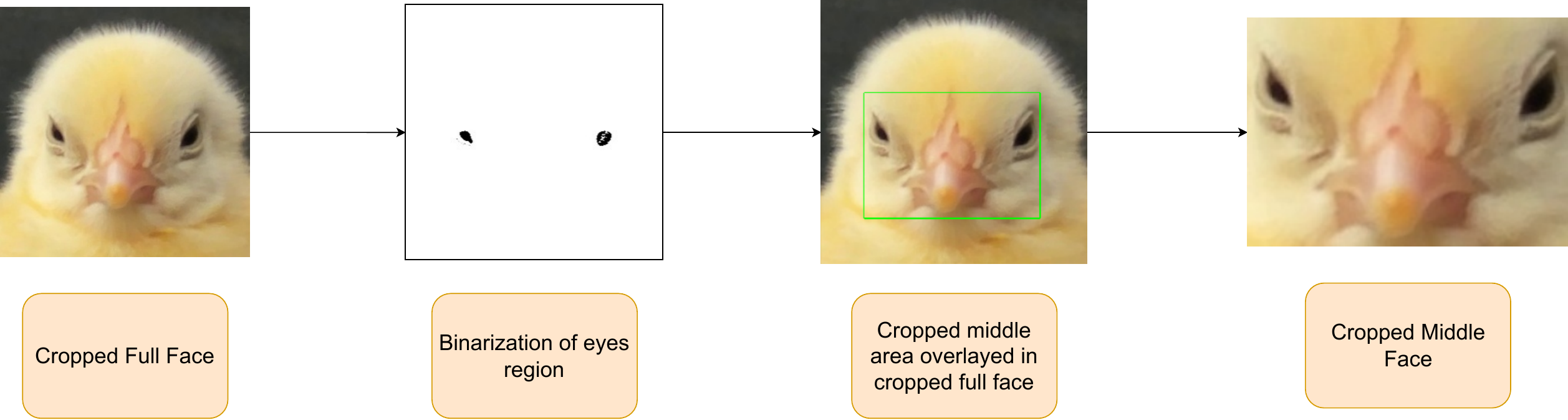}
    \caption{Overall procedure of cropping to obtain Cropped Middle Face from the Full Face.} %\hl{Include two kind of cropping feature. Choose another image which is no well aligned}
    \label{fig:cropped_middle_face}
\end{figure}

The second imagery input set aims to exclude more undesired background information and specifically put more weights on the center area of the chick's face. We utilize the Cropped Full Face images derived from the face alignment to further refine the area of interest. This refinement is achieved by employing image binarization thresholding to the grayscale image. The left and right eye key-points are then used to define a mask area, focusing the threshold to isolate the eyes region from the rest of the image \cite{kukreja2022segmentation}. The goal is to detect the leftmost pixel of the left eye and the rightmost pixel of the right eye, assuming these points define the new vertical (left and right) boundaries. The upper beak key-point is used as the horizontal top boundary. To determine the horizontal bottom boundary we identified the lowest key-point among the middle, left and right beak key-points, as it varies for each image depending on the chick's head  position. With these new boundaries, we create the new bounding box. 
%lowest key-point will depend on the image, as it can be the middle, left and right beak was used as the horizontal bottom boundary, thus creating the new bounding box.
However, to avoid losing any features with that new bounding box, we add a margin dynamically adjusted based on the distance between the left and right boundaries. To obtain this margin, we first determine the eye distance using the Euclidean distance formula. Then, we examine the space from the leftmost eye pixel to the left boundary of the Cropped Full Face image and from the rightmost eye pixel to the right boundary of the Cropped Full Face image. The eye distance is divided by half the sum of these distances and refined using a scaling factor, ensuring the margin is both proportional and symmetric around the facial features. This process, illustrated in Figure \ref{fig:cropped_middle_face},  culminates in defining a new bounding box that guarantees the conservation of the main informative facial features, which forms the Cropped Middle Face .

By using these two sets of input images, we aim to compare the results and determine which facial region provides more accurate and reliable data.

\subsection{Feature Extraction}

%In our pursuit to perform gender classification within the domain of chicks using the face aligned images, we selected several deep learning model architectures known for their high accuracy score for image classification tasks in order to extract the features: AlexNet \cite{alom2018history}, EfficientNet-B0 \cite{patel2023efficientnetb0}, Inception-V3 \cite{szegedy2015going}, two versions of ResNet \cite{he2016deep} including ResNet-50, ResNet-101, and VGG-16 \cite{simonyan2014very}.
 
In our pursuit to perform chick gender classification, we aim to extract facial features from the chicks. These characteristics are necessary to later determine the gender of the chick. To achieve this, we have selected several deep learning architectures known for their high performance in image classification tasks. The selected backbone networks include AlexNet \cite{alom2018history}, EfficientNet (i.e., EfficientNet-B0) \cite{tan2019efficientnet}, Inception (i.e., Inception-V3) \cite{szegedy2015going}, two versions of ResNet (i.e., ResNet-50, 101) \cite{he2016deep}, and VGG (i.e., VGG-16) \cite{simonyan2014very}. Each of these networks requires input images $I$ with dimensions of shape $3 \times H \times W$ (\(I \in \mathbb{R}^{3 \times H \times W}\)), representing RGB images with 3 channels. The initial layers of each backbone network extract the edges and basic features. As the network processes the image through deeper layers, it learns more abstract and complex features, outputting multi-scale feature maps. By leveraging these architectures, we aim to differentiate the gender of the chicks based on their facial features.

\subsection{Gender Classification }
 %okay
 %\hl{This should be moved to Feature Extraction}
%\hl{This is NOT classifier. All backbones listed in this session aim to extract feature.}

\begin{figure}[h]
    \centering
    \includegraphics[width=1.\textwidth]{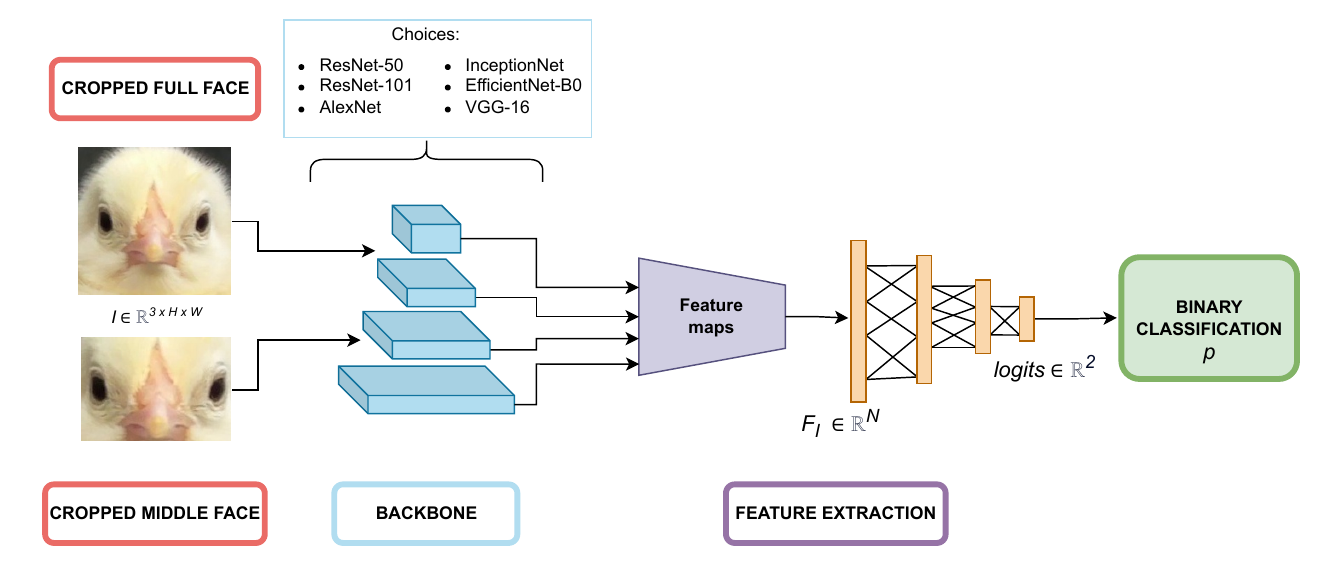}
    \caption{Neural Network Architecture For Feature extraction and classification procedure} 
    \label{fig:gender_classification}
\end{figure}

After extracting the features using various CNNs, as shown in Figure  \ref{fig:gender_classification}, the feature maps are pooled into a feature vector \(F_I \in \mathbb{R}^N \). This vector, denoted as \(F_I = f_b(I)\), where \(f_b\) represents the feature extraction process applied to the input image \(I\), is then passed through a binary classifier. We utilize these models, which are pretrained on the ImageNet dataset \cite{deng2009imagenet}, and modify the original last fully connected layer of each model with a sequence of three fully connected layers, followed by a sigmoid function. %\( z = \mathbf{W}_3 \cdot (\mathbf{W}_2 \cdot (\mathbf{W}_1 \cdot \mathbf{f} + \mathbf{b}_1) + \mathbf{b}_2) + \mathbf{b}_3 \)
The pooled feature vector \(F_I\) is transformed by the fully connected layers, denoted as \(logits = f_l(F_I)\), where \(f_l\) represents the sequence of fully connected layers. The output of these fully connected layers, referred to as logits, is then passed through the sigmoid function to produce a probability score \textit{p}. 
\begin{equation}
   p = \sigma(logits) = \frac{1}{1 + e^{-logits}}
\end{equation}

This function performs binary classification, determininig the chick's gender as either male or female by applying a classification threshold of 0.5 to the probability score \textit{p}, where \(p > 0.5\) indicates male and \(p \leq 0.5\) indicates female.

\[
\text{Gender} =
\begin{cases} 
\text{Male} & \text{if } p > 0.5 \\
\text{Female} & \text{if } p \leq 0.5
\end{cases}
\]

%The classifier in this case is a sequence of fully connected layers. The structure includes a hidden layer with 512 neurons, followed by another hidden layer with 256 neurons and final output layer with a single neuron utilizing a sigmoid function for binary classification. Adjustments to the hyperparameters of each model for our binary classification task include differentiate between male and female chicks.

%\hl{This should be moved to dataset}

\section{Experiments}
\subsection{Implementation Details}
%not fine
For benchmarking model performance, we implemented $\kappa$-fold cross-validation with $\kappa$ set to 5. We divided and iterated both sets of input images, Cropped Full Face and Cropped Middle Face images, into four folds for training and one fold for validation, organized by the individual chick's ID.

As shown in Table \ref{tab:fold_distribution}, the division across folds is balanced, not by the sheer volume of images, but by ensuring equitable distributions of gender IDs for training and validation purposes. Within each validation fold, we retrieved the best accuracy parameter over 50 epochs, facilitated by the multilayer perceptron (MLP), and averaged all accuracies obtained in each fold to determine the final performance accuracy. The models were trained using binary cross-entropy loss, Adam optimizer \cite{zhang2018improved}, and a learning rate of $1 \times 10^{-5}$.

\begin{table}[h]
\centering
\caption{Distribution of IDs and images per fold}
\label{tab:fold_distribution}
\begin{tabular}{c|c c c | c c c}
\toprule
\multirow{2}{*}{\textbf{Folds}}  & \multicolumn{2}{c}{\textbf{\# IDs}} & & \multicolumn{2}{c}{\textbf{\# Images}} \\ 
\cmidrule(lr){2-4} \cmidrule(lr){5-7}
 & \textbf{Female} & \textbf{Male} & \textbf{Total} & \textbf{Female} & \textbf{Male} & \textbf{Total} \\ 
\midrule
Fold 0 & 37 & 34 & 71 & 660 & 483 & 1,143 \\ \midrule
Fold 1 & 37 & 34 & 71 & 595 & 644 & 1,239 \\ \midrule
Fold 2 & 37 & 34 & 71 & 629 & 515 & 1,144 \\ \midrule
Fold 3 & 37 & 34 & 71 & 762 & 305 & 1,067 \\ \midrule
Fold 4 & 36 & 33 & 69 & 641 & 493 & 1,134 \\
\bottomrule
\end{tabular}

\end{table}
Note: In each cycle of a 5-fold cross-validation, 4 folds are used for training and the remainder one is used for evaluation.

\subsection{Evaluation Metrics}
To evaluate the efficiency and performance of our facial chick sexing classification model, we emphasize the use of metric scores suitable for balanced datasets: Accuracy, Precision, Recall, F1-score, and Area Under the Receiver Operating Characteristic (ROC) Curve (AUC). These criteria permits understanding the percentage of correctly classified predictions in consideration of the True Positive (TP), True Negative (TN), False Positive (FP), and False Negative (FN) values, which are obtained from the confusion matrix, represented as follows:

\begin{table}[tbh]
\centering
\renewcommand{\arraystretch}{0.8}
\setlength{\tabcolsep}{2pt}
\resizebox{0.9\columnwidth}{!}{%
\begin{tabular}{l|c|c}
\toprule
\multicolumn{0}{c|}{} & \multicolumn{1}{c|}{\textbf{Predicted Negative (Female) } } & \multicolumn{1}{c}{\textbf{Predicted Positive (Male)}} \\ \cmidrule{1-3} 
\textbf{Actual Negative (Female)}  & TN & FP \\ \midrule
\textbf{Actual Positive (Male)}  & FN & TP \\ \bottomrule
\end{tabular}}
\end{table}

These values are used to calculate the following metrics:

\begin{eqnarray}
 {Accuracy} & = &  \frac{TP + TN}{TP + TN + FP + FN} 
\end{eqnarray}

\begin{eqnarray}
 {Precision} & = &  \frac{TP}{TP + FP}
\end{eqnarray}

\begin{eqnarray}
 {Recall} & = &  \frac{TP}{TP + FN}
\end{eqnarray}

\begin{eqnarray}
 {F1-Score} & = &  \frac{TP}{TP + 0.5 * (FP + FN)}
\end{eqnarray}
%fix first line

%\hl{Need to update AUC. Here are a few references: https://developers.google.com/machine-learning/crash-course/classification/roc-and-auc }
%The AUC is a performance metric for classification models that encompasses all the points of the ROC curve. It is a representation of the true positive rate (TPR) against the false positive rate (FPR). As the classification threshold decreases, more items are classified as positive, increasing both the FP and TP. The closer the curve comes to 1, the more effective the classifier is considered to be. In essence, a high AUC reflects a model's ability to differentiate between classes with high confidence across all thresholds, making it a robust indicator of the classifier's discriminatory power and reliability.

The AUC is a performance metric for classification models that measures the two-dimensional area under the ROC curve \cite{google2024rocauc}. This curve represents the true positive rate (TPR), also known as recall, against the false positive rate (FPR) at various classification thresholds. This metric ranges from 0 to 1, indicating completely wrong predictions and 100\% correct predictions, respectively. In essence, a higher AUC score indicates a more effective classifier. AUC is scale-invariant, meaning the model tries to differentiate between instances rather than their probabilities. It is also classification-threshold-invariant, considering an overall performance across all possible thresholds for the final AUC score. However, this metric may not suit scenarios where calibrated probabilities are needed or where minimizing specific errors is crucial.

\subsection{Quantitative Results}
%fix this section
%comparison between 6 different backbone networks
% two subsets of images Cropped Full Face and Cropped middle Face ( Section \ref{sec:}
% performance is checked in various metrics, shown in Table X

This section presents a performance comparison between six different backbone networks for chick sexing based on facial images, aiming to assess the viability of this technique. As this is a pioneer study, we do not have other works for direct comparison. Therefore, we compare the results across several backbone networks: AlexNet, EfficientNet-B0, Inception-V3, ResNet-50, ResNet-101, and VGG-16. The performance of these models was tested on two types of image crops: Cropped Full Face and Cropped Middle Face.

The performance of these models was evaluated using the metrics outlined in section 5.2, and the results are detailed in the following two tables. Table \ref{tab:comparison_varios_backbones_with_folds} provides a comprehensive breakdown of the results for each metric across all folds in the 5-fold cross-validation process. Table \ref{tab:average_performance_various_backbones} summarizes these results by presenting the average performance of each backbone across all metrics.

%Since this is a pilot work, we do not have other works to compare to, we compare the results employing several backbones. This include  AlexNet, EfficientNet-B0, Inception-V3, ResNet-50, ResNet-101, and VGG-16. In this section, there are two tables in which we show This performance comparison is checked in various metrics shown in \ref{tab:comparison_varios_backbones_with_folds}

%ideas:
%- Introducir que en esta seccion se va a comparar la performance entre (no se si es mejor especificar o decir various) backbone networks.

%- No se si mencionar que como este es un proyecto piloto no hay otros resultados con los que compararlos pero podemos comparar. (o ya se sobreentiende?). 

%- la primera tabla es mas detallada, incluyendo en detalle todos los resultados de cada metrica por cada fold (del 5-fold cross validation) 

%- la segunda tabla es el average result de la tabla anterior en cada metrica, dejando un valor unico para cada metrica.

% These metrics were chosen to provide a comprehensive assessment of each model's capability to classify accurately and handle class imbalances.

\begin{table}[tbh]
\centering
\renewcommand{\arraystretch}{0.6}
\setlength{\tabcolsep}{5pt}
\caption{Performance comparison on various network backbones with five folds.}
\resizebox{0.95\columnwidth}{!}{%
\begin{tabular}{@{}l|c||ccccc||ccccc@{}}
\toprule
\multirow{2}{*}{\textbf{Backbones}} & \multirow{2}{*}{\textbf{Folds}} & \multicolumn{5}{c|}{\textbf{Cropped Full Face}} & \multicolumn{4}{c}{\textbf{Cropped Middle Face}} \\ \cmidrule(l){3-12} 
 &  & Accuracy & Precision & Recall & F1-Score & AUC & Accuracy & Precision & Recall & F1-Score & AUC \\ \toprule

\multirow{5}{*}{AlexNet} 
 & 0 & 77.08 & 77.04 & 75.52 & 75.94 & 81.68 & 79.09 & 79.15 & 79.87 & 78.98 & 85.93 \\
 & 1 & 78.45 & 78.85 & 78.18 & 78.24 & 85.05 & 78.13 & 79.63 & 78.56 & 78.00 & 86.87 \\
 & 2 & 74.65 & 74.49 & 74.70 & 74.53 & 79.92 & 75.77 & 76.75 & 74.51 & 74.76 & 86.00 \\
 & 3 & 81.35 & 77.24 & 76.52 & 76.86 & 84.54 & 82.38 & 79.57 & 75.18 & 76.81 & 85.84 \\
 & 4 & 76.46 & 76.14 & 75.68 & 75.85 & 82.78 & 71.84 & 75.44 & 75.61 & 75.51 & 82.10 \\ \midrule
\multirow{5}{*}{EfficientNet-B0} 
 & 0 & 80.14 & 80.29 & 81.03 & 80.05 & 85.59 & 80.49 & 78.00 & 80.19 & 80.08 & 88.10 \\
 & 1 & 79.82 & 80.53 & 79.50 & 79.56 & 88.60 & 78.29 & 78.36 & 78.38 & 78.29 & 85.35 \\
 & 2 & 77.36 & 77.21 & 76.90 & 77.01 & 86.04 & 73.51 & 74.70 & 72.03 & 72.15 & 81.48 \\
 & 3 & 79.38 & 74.91 & 76.61 & 75.62 & 83.85 & 69.45 & 64.40 & 61.23 & 64.87 & 71.65 \\
 & 4 & 76.01 & 75.82 & 75.01 & 75.26 & 83.19 & 73.28 & 74.82 & 74.68 & 73.28 & 83.32 \\ \midrule
\multirow{5}{*}{Inception-V3}  
 & 0 & 79.09 & 78.57 & 78.70 & 78.63 & 84.37 & 79.27 & 78.81 & 78.55 & 78.66 & 86.37 \\
 & 1 & 80.47 & 81.08 & 80.17 & 80.24 & 89.56 & 73.69 & 73.81 & 73.81 & 73.69 & 79.54 \\
 & 2 & 79.20 & 79.00 & 78.93 & 78.96 & 85.13 & 80.94 & 80.81 & 81.09 & 80.56 & 89.51 \\
 & 3 & 82.38 & 78.34 & 79.31 & 78.79 & 86.06 & 77.51 & 73.46 & 76.68 & 74.41 & 85.51 \\
 & 4 & 79.63 & 79.69 & 80.20 & 79.55 & 87.16 & 78.75 & 79.12 & 79.56 & 78.71 & 87.26 \\ \midrule
\multirow{5}{*}{ResNet-50} 
 & 0 & 83.29 & 82.89 & 83.50 & 83.06 & 89.67 & 82.59 & 82.15 & 82.20 & 82.18 & 89.33 \\
 & 1 & \textbf{85.71} & \textbf{86.02} & \textbf{85.91} & \textbf{85.71} & \textbf{93.47} & 78.69 & 80.92 & 79.21 & 78.48 & 88.60 \\
 & 2 & 79.90 & 80.68 & 78.88 & 79.22 & 89.06 & 81.12 & 82.00 & 80.12 & 80.49 & 89.15 \\
 & 3 & 82.85 & 79.12 & 82.78 & 80.36 & 89.60 & 79.10 & 76.06 & 80.84 & 76.95 & 88.03 \\
 & 4 & 78.13 & 77.83 & 78.20 & 77.92 & 85.73 & 80.78 & 80.54 & 81.00 & 80.63 & 87.40 \\ \midrule
\multirow{5}{*}{ResNet-101} 
 & 0 & 80.93 & 80.57 & 81.21 & 80.70 & 87.56 & 78.65 & 78.92 & 79.60 & 78.57 & 83.06 \\
 & 1 & 82.41 & 82.38 & 82.36 & 82.37 & 89.70 & 76.03 & 76.45 & 76.26 & 76.01 & 83.83 \\
 & 2 & 82.17 & 82.21 & 82.53 & 82.13 & 88.58 & \textbf{83.30} & \textbf{83.13} & \textbf{83.37} & \textbf{83.21} & \textbf{91.60} \\
 & 3 & 83.32 & 79.49 & 82.52 & 80.63 & 90.75 & 80.79 & 76.95 & 78.10 & 78.10 & 89.78 \\
 & 4 & 79.63 & 80.00 & 80.46 & 79.59 & 88.94 & 80.51 & 80.17 & 80.26 & 80.26 & 86.97 \\ \midrule
\multirow{5}{*}{VGG-16} 
 & 0 & 66.40 & 65.65 & 63.94 & 64.01 & 70.35 & 57.92 & 55.51 & 50.65 & 39.89 & 59.12 \\
 & 1 & 60.05 & 62.27 & 60.76 & 59.04 & 65.18 & 64.57 & 65.93 & 65.06 & 64.22 & 69.79 \\
 & 2 & 56.99 & 56.13 & 54.24 & 51.72 & 55.69 & 63.72 & 63.47 & 62.31 & 62.15 & 63.02 \\
 & 3 & 64.67 & 53.96 & 53.34 & 53.32 & 50.63 & 64.76 & 55.00 & 54.48 & 54.89 & 59.72 \\
 & 4 & 65.43 & 65.49 & 62.84 & 62.48 & 69.01 & 61.29 & 60.25 & 59.43 & 59.33 & 62.16 \\ \bottomrule
\end{tabular}}
\label{tab:comparison_varios_backbones_with_folds}
\end{table}

The consistent performance across 5-fold accuracies confirms the reliability and robustness of the models, particularly ResNet-50. 

\begin{table}[tbh]
\centering
\renewcommand{\arraystretch}{0.8}
\setlength{\tabcolsep}{2pt}
\caption{Average performance comparison of various network backbones.}
% \caption{PERFORMANCE COMPARISON ON DIFFERENT BACKBONES USING TWO INPUT IMAGES \hl{Normal font}}
\resizebox{0.9\columnwidth}{!}{%
\begin{tabular}{l||ccccc||ccccc}
\toprule
\multicolumn{1}{c|}{} & \multicolumn{5}{c|}{\textbf{Cropped Full Face}} & \multicolumn{4}{c}{\textbf{Cropped Middle Face}} \\ \cmidrule{2-11} 
\multicolumn{1}{c|}{\raisebox{1.5ex}[0pt]{\textbf{Backbones}}}& Accuracy & Precision & Recall & F1-Score & AUC & Accuracy & Precision & Recall & F1-Score & AUC \\ \midrule
AlexNet & 77.60 & 76.75 & 76.12 & 76.28 & 82.79 & 77.44 & 78.11 & 76.75 & 76.81 & 85.35 \\ 
EfficientNet-B0 & 78.54 & 77.75 & 77.81 & 77.50 & 85.45 & 74.80 & 74.06 & 73.30 & 73.73 & 81.98   \\ 
Inception-V3 & 80.15 & 79.33 & 79.46 & 79.23 & 86.46 & 78.03 & 77.20 & 77.94 & 77.00 & 85.64 \\
ResNet-50  & {\textbf{81.98}} & {\textbf{81.31}} & 
\textbf{81.85} & 81.25 & {\textbf{89.51}} & {\textbf{80.46}} & {\textbf{80.33}} & \textbf{80.67} &{\textbf{79.75}} & {\textbf{88.50}} \\
ResNet-101 & 81.69 & 80.93 & 81.82 &{\textbf{ 81.84}} & 89.10 & 79.86 & 79.12 & 80.04 & 79.23 & 87.05 \\
VGG-16 & 62.71 & 60.70 & 59.02 & 58.11 & 62.17 & 62.45 & 60.03 & 58.39 & 56.10 & 62.76 \\ \bottomrule
\end{tabular}}
\label{tab:average_performance_various_backbones}
\end{table}

From the averaged results in Table \ref{tab:average_performance_various_backbones}, ResNet-50 consistently outperforms the other models across both image crops. For the Cropped Full Face dataset, ResNet-50 registered the highest accuracy of 81.98\%, which is 0.29\% higher than ResNet-101, 1.83\% higher than Inception-V3, 3.44\% higher than EfficientNet-B0 and 4.38\% higher than AlexNet. Additionally, ResNet-50 demonstrated superior performance in precision (81.31\%), recall (81.85\%), and F1-Score (81.25\%), demonstrating an effective balance between precision and recall. ResNet-101 also delivered strong results in F1-Score (81.84\%). Although these differences between backbones may seem small, they underscore ResNet-50's robustness in feature extraction and pattern recognition capabilities, with its 25.6 million parameters contributing to its high capability for learning complex patterns. In contrast, VGG-16 shows the lowest performance with an accuracy gap of 19.27\% compared to ResNet-50. 
%precision, recall and AUC scorres for the Cropped Full Face dataset and for the Cropped Middle Face dataset it also highlights in the F1-Score. This exemplifies the depth of ResNet-50's feature extraction and pattern recognition capabilities, as deeper networks tend to excel in recognizing complex patterns, which is beneficial for distinguishing gender from facial features of chicks.

For the Cropped Middle Face dataset, ResNet-50 again leads with an accuracy of 80.46\%, outperforming ResNet-101 by 0.60\% and Inception-V3 by 2.43\%, EfficientNet-B0 by 5.66\%, and AlexNet by 3.02\%. Also, ResNet-50 maintained strong performance in precision (80.33\%), recall (80.67\%) and F1-Score (79.75\%), still consistent even with narrower crop. ResNet-101 also showed strong results but with a slight drop in all metrics. In contrast, VGG-16 performed poorly, with the lowest scores across all metrics, highlighting the importance of choosing a suitable architecture for specific tasks.

These quantitative results suggest that more complex models, such as ResNet-50 and ResNet-101, are better suited for chick sexing tasks, offering higher accuracy and reliability, while simpler models like VGG-16 may not capture the necessary features. The performance gap between the Cropped Full Face and Cropped Middle Face datasets also indicates that cropping the image may cause the model to lose attention on key areas, resulting in reduced accuracy.

\begin{figure}[h]
    \centering
    \includegraphics[width=\linewidth]{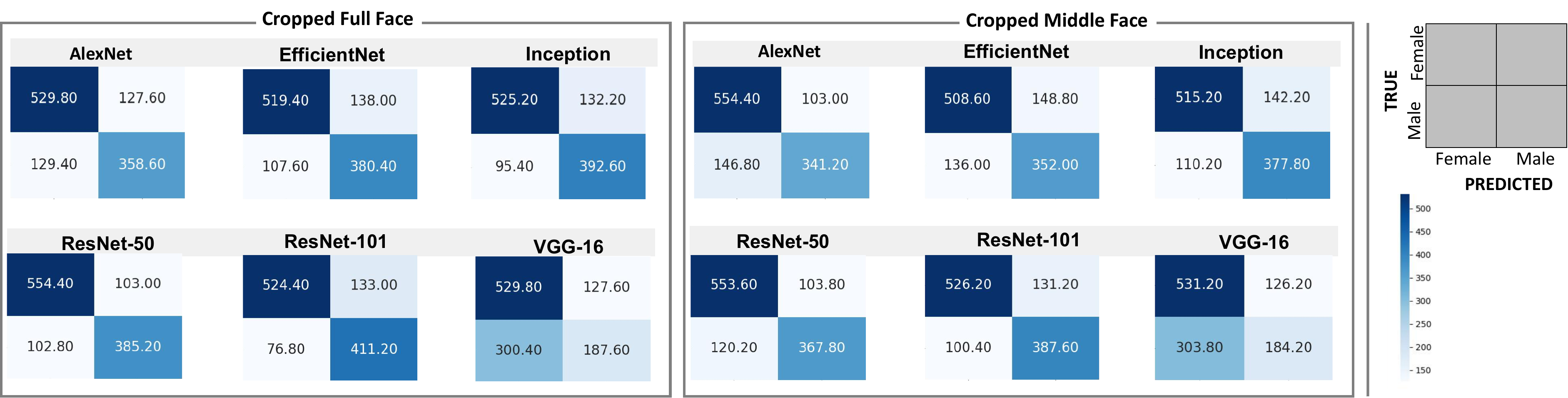}
    \caption{Averaging confusion matrices of 5 fold-cross validation on various backbones fo Cropped Full Face (left) and Cropped Middle Face (right).
    }
    \label{fig:confusion_matrices}
\end{figure}

In addition to the quantitative metrics, we also evaluate the confusion matrices for all backbone networks and both datasets to gain more insight into model performance. As shown in Figure \ref{fig:confusion_matrices}, ResNet-50 backbone reveals a more balanced distribution of incorrect classifications between males and females. This balance indicated that the model does not disproportionately favor one class over the other, ensuring fairness and reliability for chick sexing tasks. In contrast the other backbones tend to missclasify one gender more frequently than the other, resulting in worse results in the evaluated metrics. This discrepancy may also be influenced by our dataset, as it contains more female images than male images.

%"what does the amount of the confusion matrix shows?" 

%"we also have more females images and ids eventhough the dataset tries the best ob being balanced"

% \begin{figure}[h]
%     \centering
%     \includegraphics[width=0.8\linewidth]{Figs/confusion_matrices_results (2).pdf}
%     \caption{Averaging confusion matrices of 5 fold-cross validation on various backbones for both Cropped Full Face (top) and Cropped Middle Face (bottom). \hl{will be updated. include percentage}
%     }
%     \label{fig:confusion_matrices}
% \end{figure}

\subsection{Qualitative Results}
In this section, we provide a visual analysis into the decisions made by the deep learning model. To achieve this, we utilize the Grad-CAM ++ (Gradient weighted Class Activation Mapping plus plus) technique \cite{chattopadhay2018grad}, an improved extension version of the original Grad-CAM technique \cite{selvaraju2017grad}. This method generates visualizations of the model results highlight the important regions within the input image, coloring them based on where the model pays attention on. This technique calculates the gradients of the classification score related to the last convolutional layer of the model to create an importance map. This is then overlaid on top of the original image to create a feature map. These feature maps help visualize the CNN model's predictions, with warmer colors indicating areas of greater attention or importance, and cooler colors representing less influential regions on the model's output. 

\newpage
\begin{figure}[h]
    \centering
    \includegraphics[width=1.0\linewidth]{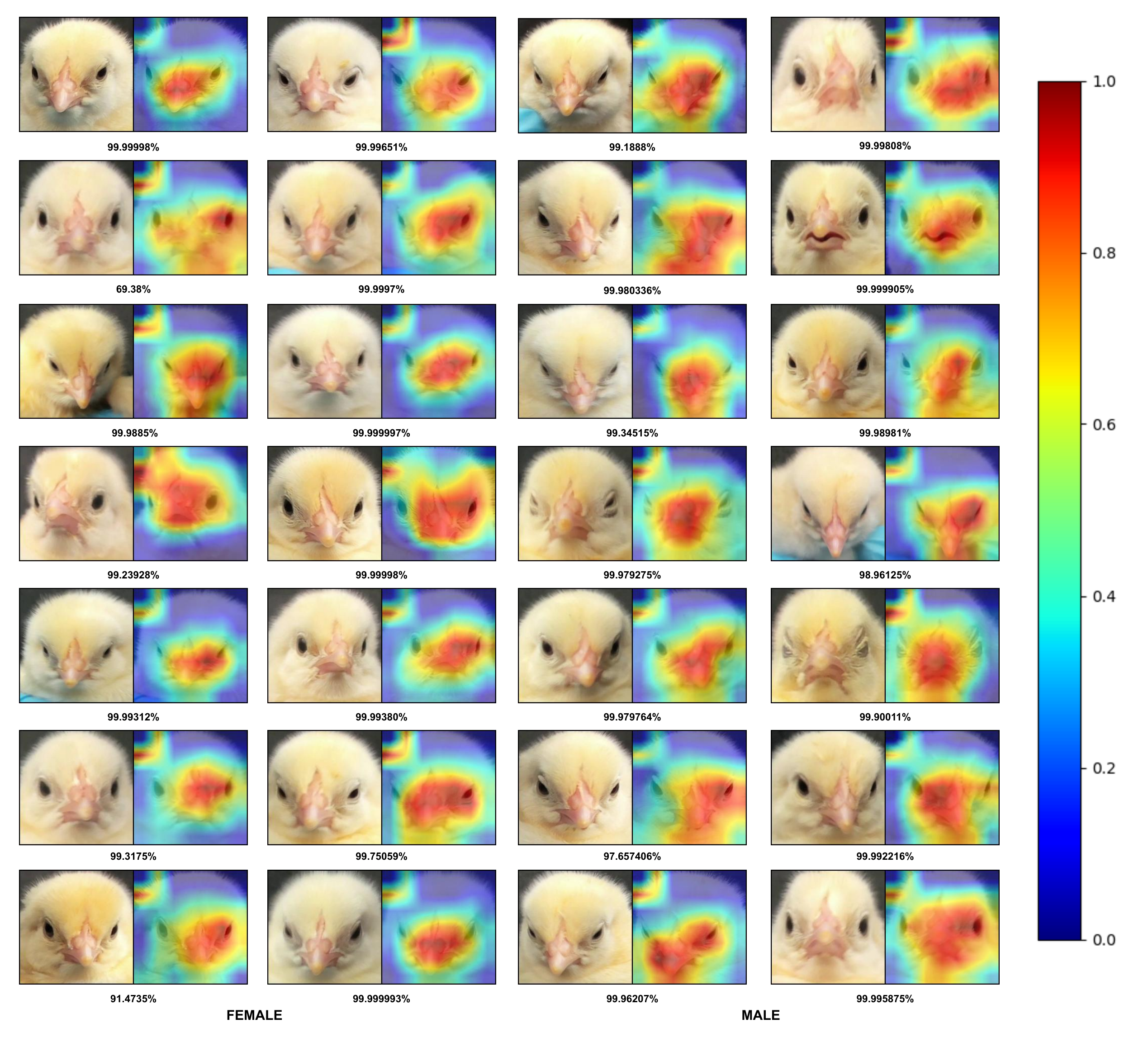}
    \caption{Feature maps generated by application of the ResNet-50 model on correctly classified Cropped Full Face images using Grad-CAMpp.% \hl{Remove the Resnet-10 on the top; Add two figures corresponding to correct classification results and incorrect classification results. Follow my instruction during the meeting and update the figures}
    }
    \label{fig:gradcam_correct_predictions_cropped_full_face}
\end{figure}

This section focuses on feature maps produced by the ResNet-50 backbone, which achieved better quantitative results than the other models. Figure \ref{fig:gradcam_correct_predictions_cropped_full_face} illustrates the impact of this technique on the classification score made by the ResNet-50 network, explaining how certain areas of the Cropped Full Face input image influence the model's output. The generated feature maps reveal that the model pays most attention to regions around the chick's beak, specifically around the middle nose key-point and the comb in warm colors (yellow and red). In contrast, areas with cooler colors, such as the upper beak region, result in less activation of the deeper layers, making them less informative to the model's decision-making process. This consistent attention aligns with the promising quantitative results obtained, ensuring that the Cropped Full Face images in Figure \ref{fig:gradcam_correct_predictions_cropped_full_face} are correctly classified with high confidence.

\begin{figure}[h]
    \centering
    \includegraphics[width=1.0\linewidth]{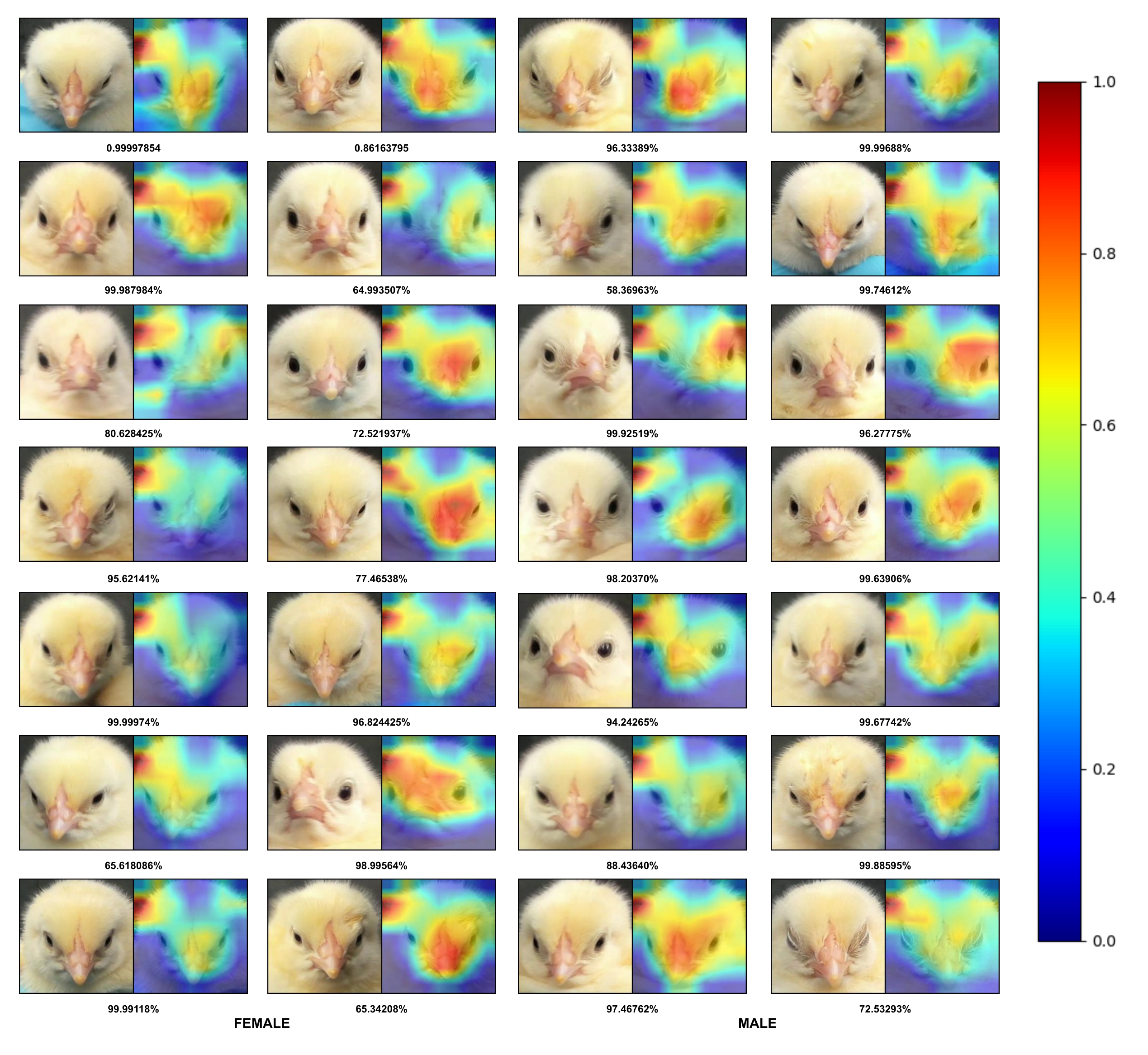}
    \caption{Feature maps generated by application of the ResNet-50 model on incorrectly classified Cropped Full Face images using Grad-CAMpp. %\hl{Remove the Resnet-10 on the top; Add two figures corresponding to correct classification results and incorrect classification results. Follow my instruction during the meeting and update the figures}
    }
    \label{fig:gradcam_incorrect_predictions_cropped_full_face}
\end{figure}

The confidence scores derived from the sigmoid function in the ResNet-50 model show high confidence in correct predictions. However, the model also shows high confidence in certain misclassifications. As shown in Figure \ref{fig:gradcam_incorrect_predictions_cropped_full_face}, some of these misclassifications, despite having high confidence scores, reveal that the model attends on more dispersed areas across the image, which likely contributes to the incorrect gender classifications. Nevertheless, there is still some attention on the regions around the beak and comb, even when the gender classifications are incorrect.

\begin{figure}[!h]
    \centering
    \includegraphics[width=1.0\linewidth]{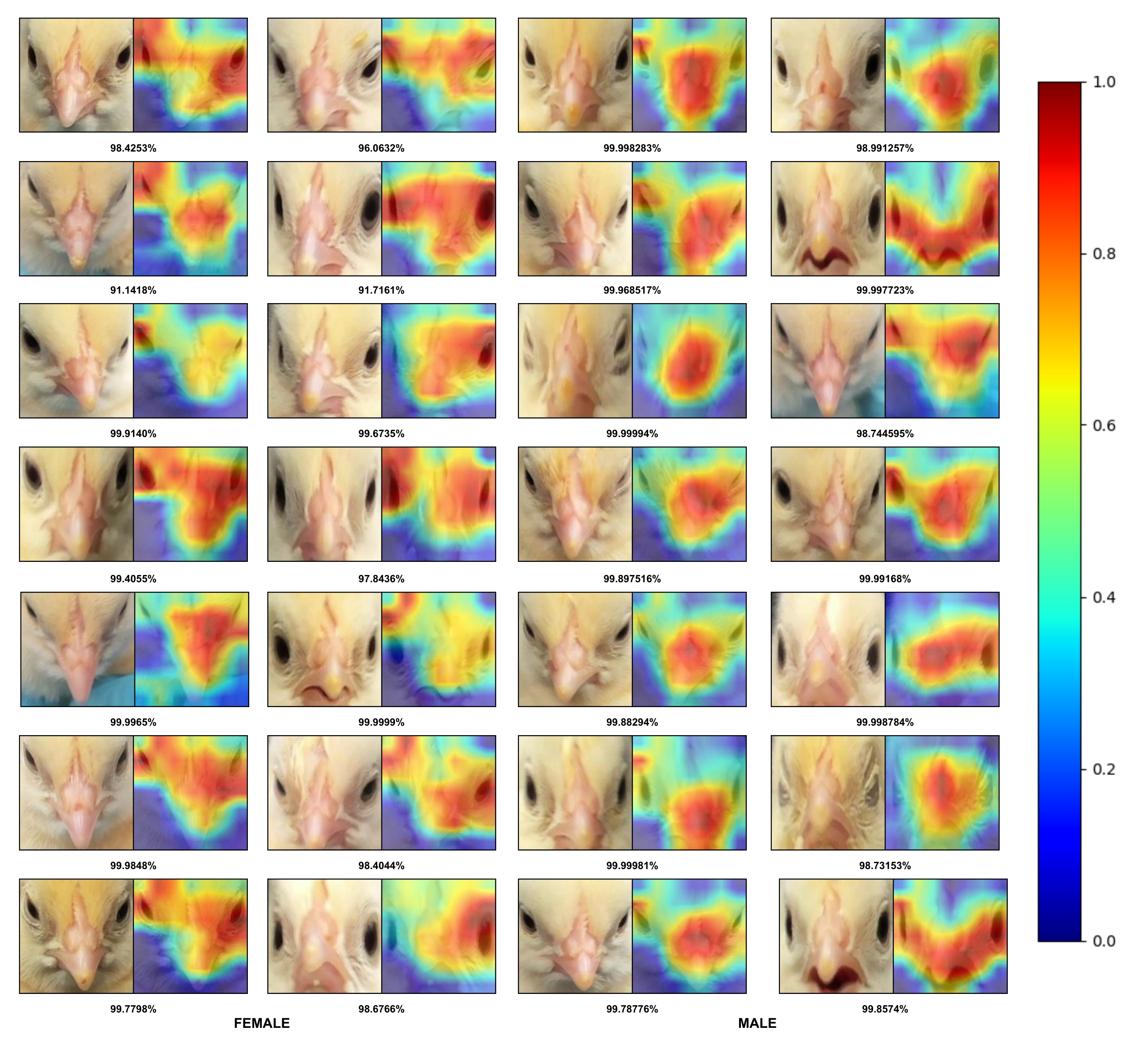}
    \caption{Feature maps generated by application of the ResNet-50 model on correctly classified Cropped Middle Face images using Grad-CAMpp.}
    \label{fig:gradcam_correct_predictions_cropped_middle_face}
\end{figure}

Similarly, Figure \ref{fig:gradcam_correct_predictions_cropped_middle_face}, shows the impact of this technique on the Cropped Middle Face images. In this figure, we observe the correct gender classification feature maps using the ResNet-50 backbone. The model is paying more attention on biased features, with a stronger emphasis on the beak area.

\begin{figure}[h]
    \centering
    \includegraphics[width=1.0\linewidth]{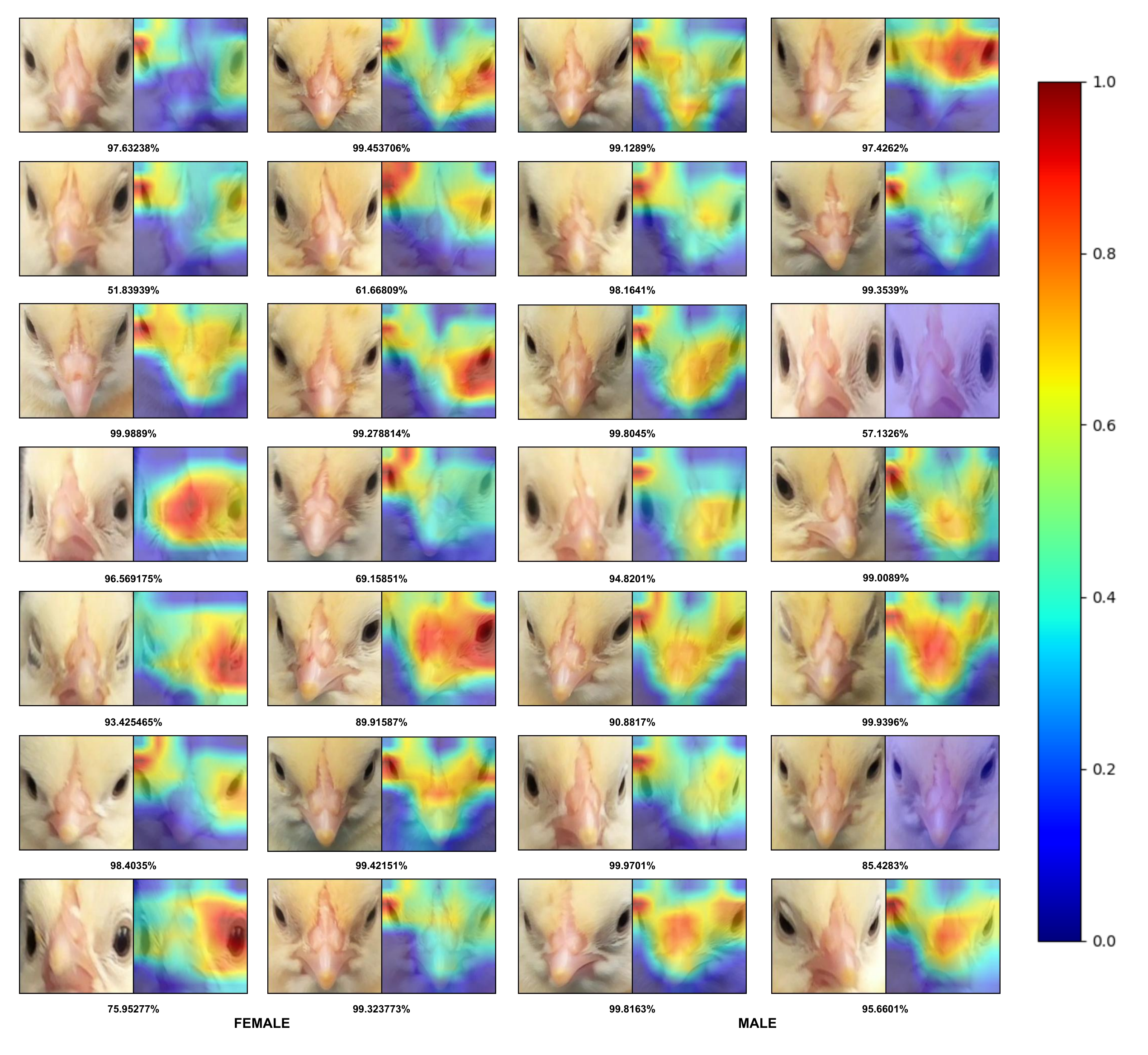}
    \caption{Feature maps generated by application of the ResNet-50 model on incorrectly classified Cropped Middle Face images using Grad-CAMpp}
    \label{fig:gradcam_incorrect_predictions_cropped_middle_face}
\end{figure}

In contrast, Figure \ref{fig:gradcam_incorrect_predictions_cropped_middle_face} shows the feature maps for incorrect classifications using the Cropped Middle Face images. In these cases, the model's attention is more dispersed, with each feature map highlighting different areas, often with cooler colors, indicating weaker confidence. These results are also reflected in the lower quantitative results obtained from the Cropped Middle Face dataset compared to the Cropped Full Face. This suggests that this cropping may lead the model to lose some key gender-differentiating feature characteristics from the chicks.

% Incorrect predictions: we can see that the model focuses on other areas like the periphery of the face. However, some of the model missclassifications still focus on the beak and comb of the chick with a high confidence score. 

%In addition to the quantitative metrics, we also analyzed the confusion matrices for each backbone and each dataset to gain further insight into the models' performance (see Figure {insert figure number here}). The confusion matrix for ResNet-50, in particular, demonstrates a balanced distribution of incorrect predictions, meaning that the number of misclassified males and females is roughly equal. This consistency indicates that ResNet-50 is not biased toward predicting one gender over the other, which enhances its reliability for chick sexing tasks.

\section{Discussion}
%\hl{Add detailed discussion on how to work valuable to poultry science and market}

In this discussion, we revisit the initial question of our study regarding the feasibility of adapting human facial recognition techniques for one-day-old chick sexing. Our findings indicate that this approach is not only viable but also holds substantial potential for industrial application in the field of chick sexing. The performance of our system is promising and suggests the possibility of opening a new market in the poultry science industry. One practical way to integrate this system for future applications is by capturing the frontal faces of the chicks during the vaccination process. Given that vaccination machines already position the chicks by holding them by the head, leaving them in a completely frontal position, installing a camera system within this machinery could enable simultaneous vaccination and facial chick sexing processes, streamlining operations and enhancing efficiency.

Despite the promising results, several limitations were encountered. The dataset used for training and validation was not extensive. A larger dataset, along with more diverse samples, could improve the robustness of the model. Additionally, the losses derived from the model's performance may require further tuning of hyperparameters. Variations in lightning, background factors, and chick stillness can influence the accuracy of facial recognition techniques. While the proposed discussion regarding the integration with vaccination and imaging processes, image quality in a high-speed machine, as well as real-time processing capabilities, need to be addressed to ensure smooth and efficient processes. By addressing these limitations, future research can further enhance the effectiveness and portability of facial chick sexing procedures benefiting animal welfare.

\section{Conclusion}

In conclusion, in this research we explored the application of human facial recognition technologies adapted for chick sexing, highlighting the potential of computer vision techniques. The development of a multi-view recording system and a semi-automated annotation process has enabled efficient handling of variations in chick frontal or near frontal cases, contributing significantly to the accuracy of gender classification. The exploration of human facial recognition technologies adapted to chicks has shown promising results, especially with the application of deep learning models using ResNet-50. Additionally, qualitative insights from Grad-CAM++ visualizations confirm that the model consistently retrieves on key anatomical features, such as the chick's beak and comb, ensuring the reliability of its predictions. However, visual analysis also suggests that our cropping strategy may lead to information loss or incorrect attention, affecting performance. This approach not only offers a non-invasive alternative to traditional methods but also aligns with the need for innovative solutions in the poultry science industry, reducing labor and improving animal welfare.

%\newpage

\section{Acknowledgments}

This material is based upon work supported by Cobb Vantress Inc. 
% %the National Science Foundation (NSF) under Award No OIA-1946391 RII Track-1, NSF 2119691 AI SUSTEIN.
% NSF 1920920 RII Track 2 FEC, NSF 2223793 EFRI BRAID, NSF 2119691 AI SUSTEIN, NSF 2236302, and the National Institutes of Health (NIH) 1R01CA277739-01.

\bibliographystyle{elsarticle-harv} 
\bibliography{Refs}

\begin{thebibliography}{45}
\expandafter\ifx\csname natexlab\endcsname\relax\def\natexlab#1{#1}\fi
\providecommand{\url}[1]{\texttt{#1}}
\providecommand{\href}[2]{#2}
\providecommand{\path}[1]{#1}
\providecommand{\DOIprefix}{doi:}
\providecommand{\ArXivprefix}{arXiv:}
\providecommand{\URLprefix}{URL: }
\providecommand{\Pubmedprefix}{pmid:}
\providecommand{\doi}[1]{\href{http://dx.doi.org/#1}{\path{#1}}}
\providecommand{\Pubmed}[1]{\href{pmid:#1}{\path{#1}}}
\providecommand{\bibinfo}[2]{#2}
\ifx\xfnm\relax \def\xfnm[#1]{\unskip,\space#1}\fi
%Type = Article
\bibitem[{Abdullah et~al.(2010)Abdullah, Al-Beitawi, Rjoup, Qudsieh and Ishmais}]{abdullah2010growth}
\bibinfo{author}{Abdullah, A.Y.}, \bibinfo{author}{Al-Beitawi, N.A.}, \bibinfo{author}{Rjoup, M.M.}, \bibinfo{author}{Qudsieh, R.I.}, \bibinfo{author}{Ishmais, M.A.}, \bibinfo{year}{2010}.
\newblock \bibinfo{title}{Growth performance, carcass and meat quality characteristics of different commercial crosses of broiler strains of chicken}.
\newblock \bibinfo{journal}{The journal of poultry science} \bibinfo{volume}{47}, \bibinfo{pages}{13--21}.
%Type = Article
\bibitem[{Afifi and Abdelhamed(2019)}]{afifi2019afif4}
\bibinfo{author}{Afifi, M.}, \bibinfo{author}{Abdelhamed, A.}, \bibinfo{year}{2019}.
\newblock \bibinfo{title}{Afif4: Deep gender classification based on adaboost-based fusion of isolated facial features and foggy faces}.
\newblock \bibinfo{journal}{Journal of Visual Communication and Image Representation} \bibinfo{volume}{62}, \bibinfo{pages}{77--86}.
%Type = Article
\bibitem[{Alin et~al.(2019)Alin, Fujitani, Kashimori, Suzuki, Ogawa and Kondo}]{alin2019non}
\bibinfo{author}{Alin, K.}, \bibinfo{author}{Fujitani, S.}, \bibinfo{author}{Kashimori, A.}, \bibinfo{author}{Suzuki, T.}, \bibinfo{author}{Ogawa, Y.}, \bibinfo{author}{Kondo, N.}, \bibinfo{year}{2019}.
\newblock \bibinfo{title}{Non-invasive broiler chick embryo sexing based on opacity value of incubated eggs}.
\newblock \bibinfo{journal}{Computers and electronics in agriculture} \bibinfo{volume}{158}, \bibinfo{pages}{30--35}.
%Type = Article
\bibitem[{Alom et~al.(2018)Alom, Taha, Yakopcic, Westberg, Sidike, Nasrin, Van~Esesn, Awwal and Asari}]{alom2018history}
\bibinfo{author}{Alom, M.Z.}, \bibinfo{author}{Taha, T.M.}, \bibinfo{author}{Yakopcic, C.}, \bibinfo{author}{Westberg, S.}, \bibinfo{author}{Sidike, P.}, \bibinfo{author}{Nasrin, M.S.}, \bibinfo{author}{Van~Esesn, B.C.}, \bibinfo{author}{Awwal, A.A.S.}, \bibinfo{author}{Asari, V.K.}, \bibinfo{year}{2018}.
\newblock \bibinfo{title}{The history began from alexnet: A comprehensive survey on deep learning approaches}.
\newblock \bibinfo{journal}{arXiv preprint arXiv:1803.01164} .
%Type = Article
\bibitem[{Antipov et~al.(2016)Antipov, Berrani and Dugelay}]{antipov2016minimalistic}
\bibinfo{author}{Antipov, G.}, \bibinfo{author}{Berrani, S.A.}, \bibinfo{author}{Dugelay, J.L.}, \bibinfo{year}{2016}.
\newblock \bibinfo{title}{Minimalistic cnn-based ensemble model for gender prediction from face images}.
\newblock \bibinfo{journal}{Pattern recognition letters} \bibinfo{volume}{70}, \bibinfo{pages}{59--65}.
%Type = Article
\bibitem[{Biederman and Shiffrar(1987)}]{biederman1987sexing}
\bibinfo{author}{Biederman, I.}, \bibinfo{author}{Shiffrar, M.M.}, \bibinfo{year}{1987}.
\newblock \bibinfo{title}{Sexing day-old chicks: A case study and expert systems analysis of a difficult perceptual-learning task.}
\newblock \bibinfo{journal}{Journal of Experimental Psychology: Learning, memory, and cognition} \bibinfo{volume}{13}, \bibinfo{pages}{640}.
%Type = Article
\bibitem[{Buchala et~al.(2005)Buchala, Davey, Frank, Loomes and Gale}]{buchala2005role}
\bibinfo{author}{Buchala, S.}, \bibinfo{author}{Davey, N.}, \bibinfo{author}{Frank, R.J.}, \bibinfo{author}{Loomes, M.}, \bibinfo{author}{Gale, T.M.}, \bibinfo{year}{2005}.
\newblock \bibinfo{title}{The role of global and feature based information in gender classification of faces: a comparison of human performance and computational models}.
\newblock \bibinfo{journal}{International Journal of Neural Systems} \bibinfo{volume}{15}, \bibinfo{pages}{121--128}.
%Type = Inproceedings
\bibitem[{Chattopadhay et~al.(2018)Chattopadhay, Sarkar, Howlader and Balasubramanian}]{chattopadhay2018grad}
\bibinfo{author}{Chattopadhay, A.}, \bibinfo{author}{Sarkar, A.}, \bibinfo{author}{Howlader, P.}, \bibinfo{author}{Balasubramanian, V.N.}, \bibinfo{year}{2018}.
\newblock \bibinfo{title}{Grad-cam++: Generalized gradient-based visual explanations for deep convolutional networks}, in: \bibinfo{booktitle}{2018 IEEE winter conference on applications of computer vision (WACV)}, \bibinfo{organization}{IEEE}. pp. \bibinfo{pages}{839--847}.
%Type = Article
\bibitem[{Dakpogan et~al.(2012)Dakpogan, Salifou, Aboh and Chrysostome}]{dakpogan2012effectiveness}
\bibinfo{author}{Dakpogan, H.B.}, \bibinfo{author}{Salifou, S.}, \bibinfo{author}{Aboh, A.}, \bibinfo{author}{Chrysostome, C.}, \bibinfo{year}{2012}.
\newblock \bibinfo{title}{Effectiveness of a sexing technique on free-range day-old chick.}
\newblock \bibinfo{journal}{Journal of Animal and Plant Sciences (JAPS)} \bibinfo{volume}{16}, \bibinfo{pages}{2336--2342}.
%Type = Inproceedings
\bibitem[{Deng et~al.(2009)Deng, Dong, Socher, Li, Li and Fei-Fei}]{deng2009imagenet}
\bibinfo{author}{Deng, J.}, \bibinfo{author}{Dong, W.}, \bibinfo{author}{Socher, R.}, \bibinfo{author}{Li, L.J.}, \bibinfo{author}{Li, K.}, \bibinfo{author}{Fei-Fei, L.}, \bibinfo{year}{2009}.
\newblock \bibinfo{title}{Imagenet: A large-scale hierarchical image database}, in: \bibinfo{booktitle}{2009 IEEE conference on computer vision and pattern recognition}, \bibinfo{organization}{Ieee}. pp. \bibinfo{pages}{248--255}.
%Type = Inproceedings
\bibitem[{D’Amelio et~al.(2019)D’Amelio, Cuculo and Bursic}]{d2019gender}
\bibinfo{author}{D’Amelio, A.}, \bibinfo{author}{Cuculo, V.}, \bibinfo{author}{Bursic, S.}, \bibinfo{year}{2019}.
\newblock \bibinfo{title}{Gender recognition in the wild with small sample size-a dictionary learning approach}, in: \bibinfo{booktitle}{International Symposium on Formal Methods}, \bibinfo{organization}{Springer}. pp. \bibinfo{pages}{162--169}.
%Type = Article
\bibitem[{England et~al.(2023)England, Gharib-Naseri, Kheravii and Wu}]{england2023influence}
\bibinfo{author}{England, A.}, \bibinfo{author}{Gharib-Naseri, K.}, \bibinfo{author}{Kheravii, S.K.}, \bibinfo{author}{Wu, S.B.}, \bibinfo{year}{2023}.
\newblock \bibinfo{title}{Influence of sex and rearing method on performance and flock uniformity in broilers—implications for research settings}.
\newblock \bibinfo{journal}{Animal Nutrition} \bibinfo{volume}{12}, \bibinfo{pages}{276--283}.
%Type = Article
\bibitem[{England et~al.(2021)England, Kheravii, Musigwa, Kumar, Daneshmand, Sharma, Gharib-Naseri and Wu}]{england2021sexing}
\bibinfo{author}{England, A.}, \bibinfo{author}{Kheravii, S.}, \bibinfo{author}{Musigwa, S.}, \bibinfo{author}{Kumar, A.}, \bibinfo{author}{Daneshmand, A.}, \bibinfo{author}{Sharma, N.}, \bibinfo{author}{Gharib-Naseri, K.}, \bibinfo{author}{Wu, S.}, \bibinfo{year}{2021}.
\newblock \bibinfo{title}{Sexing chickens (gallus gallus domesticus) with high-resolution melting analysis using feather crude dna}.
\newblock \bibinfo{journal}{Poultry Science} \bibinfo{volume}{100}, \bibinfo{pages}{100924}.
%Type = Misc
\bibitem[{{Google Developers}(2024)}]{google2024rocauc}
\bibinfo{author}{{Google Developers}}, \bibinfo{year}{2024}.
\newblock \bibinfo{title}{Roc and auc}.
\newblock \bibinfo{howpublished}{\url{https://developers.google.com/machine-learning/crash-course/classification/roc-and-auc}}.
\newblock \bibinfo{note}{Accessed: Sep. 2024}.
%Type = Article
\bibitem[{Gupta and Nain(2023)}]{gupta2023single}
\bibinfo{author}{Gupta, S.K.}, \bibinfo{author}{Nain, N.}, \bibinfo{year}{2023}.
\newblock \bibinfo{title}{Single attribute and multi attribute facial gender and age estimation}.
\newblock \bibinfo{journal}{Multimedia Tools and Applications} \bibinfo{volume}{82}, \bibinfo{pages}{1289--1311}.
%Type = Inproceedings
\bibitem[{He et~al.(2016)He, Zhang, Ren and Sun}]{he2016deep}
\bibinfo{author}{He, K.}, \bibinfo{author}{Zhang, X.}, \bibinfo{author}{Ren, S.}, \bibinfo{author}{Sun, J.}, \bibinfo{year}{2016}.
\newblock \bibinfo{title}{Deep residual learning for image recognition}, in: \bibinfo{booktitle}{Proceedings of the IEEE conference on computer vision and pattern recognition}, pp. \bibinfo{pages}{770--778}.
%Type = Article
\bibitem[{He et~al.(2019)He, Martins, Huguenin, Van, Manso, Galindo, Gregoire, Catherinot, Molina and Espeut}]{he2019simple}
\bibinfo{author}{He, L.}, \bibinfo{author}{Martins, P.}, \bibinfo{author}{Huguenin, J.}, \bibinfo{author}{Van, T.N.N.}, \bibinfo{author}{Manso, T.}, \bibinfo{author}{Galindo, T.}, \bibinfo{author}{Gregoire, F.}, \bibinfo{author}{Catherinot, L.}, \bibinfo{author}{Molina, F.}, \bibinfo{author}{Espeut, J.}, \bibinfo{year}{2019}.
\newblock \bibinfo{title}{Simple, sensitive and robust chicken specific sexing assays, compliant with large scale analysis}.
\newblock \bibinfo{journal}{PLoS One} \bibinfo{volume}{14}, \bibinfo{pages}{e0213033}.
%Type = Article
\bibitem[{Iswati et~al.(2020)Iswati, Natsir, Ciptadi and Susilawati}]{iswati2020sexing}
\bibinfo{author}{Iswati}, \bibinfo{author}{Natsir, M.H.}, \bibinfo{author}{Ciptadi, G.}, \bibinfo{author}{Susilawati, T.}, \bibinfo{year}{2020}.
\newblock \bibinfo{title}{Sexing on a day-old chick of gold arabic chicken (gallus turcicus) using feather sexing method.}
\newblock \bibinfo{journal}{Journal of Animal and Plant Sciences (JAPS)} \bibinfo{volume}{44}, \bibinfo{pages}{7708--7716}.
%Type = Article
\bibitem[{Jia et~al.(2023)Jia, Li, Zhu, Wang, Zhao and Zhao}]{jia2023review}
\bibinfo{author}{Jia, N.}, \bibinfo{author}{Li, B.}, \bibinfo{author}{Zhu, J.}, \bibinfo{author}{Wang, H.}, \bibinfo{author}{Zhao, Y.}, \bibinfo{author}{Zhao, W.}, \bibinfo{year}{2023}.
\newblock \bibinfo{title}{A review of key techniques for in ovo sexing of chicken eggs}.
\newblock \bibinfo{journal}{Agriculture} \bibinfo{volume}{13}, \bibinfo{pages}{677}.
%Type = Inproceedings
\bibitem[{Jia et~al.(2016)Jia, Lansdall-Welfare and Cristianini}]{jia2016gender}
\bibinfo{author}{Jia, S.}, \bibinfo{author}{Lansdall-Welfare, T.}, \bibinfo{author}{Cristianini, N.}, \bibinfo{year}{2016}.
\newblock \bibinfo{title}{Gender classification by deep learning on millions of weakly labelled images}, in: \bibinfo{booktitle}{2016 IEEE 16th International Conference on Data Mining Workshops (ICDMW)}, \bibinfo{organization}{IEEE}. pp. \bibinfo{pages}{462--467}.
%Type = Article
\bibitem[{Jocher et~al.(2022)Jocher, Chaurasia, Stoken, Borovec, Kwon, Michael, Fang, Yifu, Wong, Montes et~al.}]{jocher2022ultralytics}
\bibinfo{author}{Jocher, G.}, \bibinfo{author}{Chaurasia, A.}, \bibinfo{author}{Stoken, A.}, \bibinfo{author}{Borovec, J.}, \bibinfo{author}{Kwon, Y.}, \bibinfo{author}{Michael, K.}, \bibinfo{author}{Fang, J.}, \bibinfo{author}{Yifu, Z.}, \bibinfo{author}{Wong, C.}, \bibinfo{author}{Montes, D.}, et~al., \bibinfo{year}{2022}.
\newblock \bibinfo{title}{ultralytics/yolov5: v7. 0-yolov5 sota realtime instance segmentation}.
\newblock \bibinfo{journal}{Zenodo} .
%Type = Article
\bibitem[{Jones et~al.(1991)Jones, Shearer and Gates}]{jones1991edge}
\bibinfo{author}{Jones, P.}, \bibinfo{author}{Shearer, S.}, \bibinfo{author}{Gates, R.}, \bibinfo{year}{1991}.
\newblock \bibinfo{title}{Edge extraction algorithm for feather sexing poultry chicks}.
\newblock \bibinfo{journal}{Transactions of the ASAE} \bibinfo{volume}{34}, \bibinfo{pages}{635--0640}.
%Type = Article
\bibitem[{Kalejaiye-Matti(2021)}]{kalejaiye2021poultry}
\bibinfo{author}{Kalejaiye-Matti, R.}, \bibinfo{year}{2021}.
\newblock \bibinfo{title}{Poultry production}.
\newblock \bibinfo{journal}{Agricultural Technology for Colleges} , \bibinfo{pages}{281}.
%Type = Article
\bibitem[{Khan et~al.(2019)Khan, Attique, Syed and Gul}]{khan2019automatic}
\bibinfo{author}{Khan, K.}, \bibinfo{author}{Attique, M.}, \bibinfo{author}{Syed, I.}, \bibinfo{author}{Gul, A.}, \bibinfo{year}{2019}.
\newblock \bibinfo{title}{Automatic gender classification through face segmentation}.
\newblock \bibinfo{journal}{Symmetry} \bibinfo{volume}{11}, \bibinfo{pages}{770}.
%Type = Inproceedings
\bibitem[{Kukreja et~al.(2022)}]{kukreja2022segmentation}
\bibinfo{author}{Kukreja, V.}, et~al., \bibinfo{year}{2022}.
\newblock \bibinfo{title}{Segmentation and contour detection for handwritten mathematical expressions using opencv}, in: \bibinfo{booktitle}{2022 international conference on decision aid sciences and applications (DASA)}, \bibinfo{organization}{IEEE}. pp. \bibinfo{pages}{305--310}.
%Type = Inproceedings
\bibitem[{Lee et~al.(2013)Lee, Lin and Huang}]{lee2013novel}
\bibinfo{author}{Lee, J.D.}, \bibinfo{author}{Lin, C.Y.}, \bibinfo{author}{Huang, C.H.}, \bibinfo{year}{2013}.
\newblock \bibinfo{title}{Novel features selection for gender classification}, in: \bibinfo{booktitle}{2013 IEEE International Conference on Mechatronics and Automation}, \bibinfo{organization}{IEEE}. pp. \bibinfo{pages}{785--790}.
%Type = Inproceedings
\bibitem[{Lee et~al.(2010)Lee, Hung and Hung}]{lee2010automatic}
\bibinfo{author}{Lee, P.H.}, \bibinfo{author}{Hung, J.Y.}, \bibinfo{author}{Hung, Y.P.}, \bibinfo{year}{2010}.
\newblock \bibinfo{title}{Automatic gender recognition using fusion of facial strips}, in: \bibinfo{booktitle}{2010 20th International Conference on Pattern Recognition}, \bibinfo{organization}{IEEE}. pp. \bibinfo{pages}{1140--1143}.
%Type = Inproceedings
\bibitem[{Leng and Wang(2008)}]{leng2008improving}
\bibinfo{author}{Leng, X.}, \bibinfo{author}{Wang, Y.}, \bibinfo{year}{2008}.
\newblock \bibinfo{title}{Improving generalization for gender classification}, in: \bibinfo{booktitle}{2008 15th IEEE International Conference on Image Processing}, \bibinfo{organization}{IEEE}. pp. \bibinfo{pages}{1656--1659}.
%Type = Article
\bibitem[{Li et~al.(2012)Li, Lian and Lu}]{li2012gender}
\bibinfo{author}{Li, B.}, \bibinfo{author}{Lian, X.C.}, \bibinfo{author}{Lu, B.L.}, \bibinfo{year}{2012}.
\newblock \bibinfo{title}{Gender classification by combining clothing, hair and facial component classifiers}.
\newblock \bibinfo{journal}{Neurocomputing} \bibinfo{volume}{76}, \bibinfo{pages}{18--27}.
%Type = Inproceedings
\bibitem[{Lin et~al.(2014)Lin, Maire, Belongie, Hays, Perona, Ramanan, Doll{\'a}r and Zitnick}]{lin2014microsoft}
\bibinfo{author}{Lin, T.Y.}, \bibinfo{author}{Maire, M.}, \bibinfo{author}{Belongie, S.}, \bibinfo{author}{Hays, J.}, \bibinfo{author}{Perona, P.}, \bibinfo{author}{Ramanan, D.}, \bibinfo{author}{Doll{\'a}r, P.}, \bibinfo{author}{Zitnick, C.L.}, \bibinfo{year}{2014}.
\newblock \bibinfo{title}{Microsoft coco: Common objects in context}, in: \bibinfo{booktitle}{Computer Vision--ECCV 2014: 13th European Conference, Zurich, Switzerland, September 6-12, 2014, Proceedings, Part V 13}, \bibinfo{organization}{Springer}. pp. \bibinfo{pages}{740--755}.
%Type = Article
\bibitem[{Nandi et~al.(2003)Nandi, McBride, Blanco and Clinton}]{nandi2003sex}
\bibinfo{author}{Nandi, S.}, \bibinfo{author}{McBride, D.}, \bibinfo{author}{Blanco, R.}, \bibinfo{author}{Clinton, M.}, \bibinfo{year}{2003}.
\newblock \bibinfo{title}{Sex diagnosis and sex determination}.
\newblock \bibinfo{journal}{World's Poultry Science Journal} \bibinfo{volume}{59}, \bibinfo{pages}{8--14}.
%Type = Article
\bibitem[{Phelps et~al.(2003)Phelps, Bhutada, Bryan, Chalker, Ferrell, Neuman, Ricks, Tran and Butt}]{phelps2003automated}
\bibinfo{author}{Phelps, P.}, \bibinfo{author}{Bhutada, A.}, \bibinfo{author}{Bryan, S.}, \bibinfo{author}{Chalker, A.}, \bibinfo{author}{Ferrell, B.}, \bibinfo{author}{Neuman, S.}, \bibinfo{author}{Ricks, C.}, \bibinfo{author}{Tran, H.}, \bibinfo{author}{Butt, T.}, \bibinfo{year}{2003}.
\newblock \bibinfo{title}{Automated identification of male layer chicks prior to hatch}.
\newblock \bibinfo{journal}{Worlds Poult Sci J} \bibinfo{volume}{59}, \bibinfo{pages}{33--8}.
%Type = Inproceedings
\bibitem[{Selvaraju et~al.(2017)Selvaraju, Cogswell, Das, Vedantam, Parikh and Batra}]{selvaraju2017grad}
\bibinfo{author}{Selvaraju, R.R.}, \bibinfo{author}{Cogswell, M.}, \bibinfo{author}{Das, A.}, \bibinfo{author}{Vedantam, R.}, \bibinfo{author}{Parikh, D.}, \bibinfo{author}{Batra, D.}, \bibinfo{year}{2017}.
\newblock \bibinfo{title}{Grad-cam: Visual explanations from deep networks via gradient-based localization}, in: \bibinfo{booktitle}{Proceedings of the IEEE international conference on computer vision}, pp. \bibinfo{pages}{618--626}.
%Type = Inproceedings
\bibitem[{Simanjuntak and Azzopardi(2020)}]{simanjuntak2020fusion}
\bibinfo{author}{Simanjuntak, F.}, \bibinfo{author}{Azzopardi, G.}, \bibinfo{year}{2020}.
\newblock \bibinfo{title}{Fusion of cnn-and cosfire-based features with application to gender recognition from face images}, in: \bibinfo{booktitle}{Advances in Computer Vision: Proceedings of the 2019 Computer Vision Conference (CVC), Volume 1 1}, \bibinfo{organization}{Springer}. pp. \bibinfo{pages}{444--458}.
%Type = Article
\bibitem[{Simonyan and Zisserman(2014)}]{simonyan2014very}
\bibinfo{author}{Simonyan, K.}, \bibinfo{author}{Zisserman, A.}, \bibinfo{year}{2014}.
\newblock \bibinfo{title}{Very deep convolutional networks for large-scale image recognition}.
\newblock \bibinfo{journal}{arXiv preprint arXiv:1409.1556} .
%Type = Inproceedings
\bibitem[{Soliman-Cuevas and Linsangan(2023)}]{soliman2023day}
\bibinfo{author}{Soliman-Cuevas, H.L.}, \bibinfo{author}{Linsangan, N.B.}, \bibinfo{year}{2023}.
\newblock \bibinfo{title}{Day-old chick sexing using convolutional neural network (cnn) and computer vision}, in: \bibinfo{booktitle}{2023 IEEE International Conference on Artificial Intelligence in Engineering and Technology (IICAIET)}, \bibinfo{organization}{IEEE}. pp. \bibinfo{pages}{45--49}.
%Type = Inproceedings
\bibitem[{Sumi et~al.(2021)Sumi, Hossain, Islam and Andersson}]{sumi2021human}
\bibinfo{author}{Sumi, T.A.}, \bibinfo{author}{Hossain, M.S.}, \bibinfo{author}{Islam, R.U.}, \bibinfo{author}{Andersson, K.}, \bibinfo{year}{2021}.
\newblock \bibinfo{title}{Human gender detection from facial images using convolution neural network}, in: \bibinfo{booktitle}{Applied Intelligence and Informatics: First International Conference, AII 2021, Nottingham, UK, July 30--31, 2021, Proceedings 1}, \bibinfo{organization}{Springer}. pp. \bibinfo{pages}{188--203}.
%Type = Inproceedings
\bibitem[{Szegedy et~al.(2015)Szegedy, Liu, Jia, Sermanet, Reed, Anguelov, Erhan, Vanhoucke and Rabinovich}]{szegedy2015going}
\bibinfo{author}{Szegedy, C.}, \bibinfo{author}{Liu, W.}, \bibinfo{author}{Jia, Y.}, \bibinfo{author}{Sermanet, P.}, \bibinfo{author}{Reed, S.}, \bibinfo{author}{Anguelov, D.}, \bibinfo{author}{Erhan, D.}, \bibinfo{author}{Vanhoucke, V.}, \bibinfo{author}{Rabinovich, A.}, \bibinfo{year}{2015}.
\newblock \bibinfo{title}{Going deeper with convolutions}, in: \bibinfo{booktitle}{Proceedings of the IEEE conference on computer vision and pattern recognition}, pp. \bibinfo{pages}{1--9}.
%Type = Inproceedings
\bibitem[{Tan and Le(2019)}]{tan2019efficientnet}
\bibinfo{author}{Tan, M.}, \bibinfo{author}{Le, Q.}, \bibinfo{year}{2019}.
\newblock \bibinfo{title}{Efficientnet: Rethinking model scaling for convolutional neural networks}, in: \bibinfo{booktitle}{International conference on machine learning}, \bibinfo{organization}{PMLR}. pp. \bibinfo{pages}{6105--6114}.
%Type = Misc
\bibitem[{Wada(2016)}]{labelme2016}
\bibinfo{author}{Wada, K.}, \bibinfo{year}{2016}.
\newblock \bibinfo{title}{{labelme: Image Polygonal Annotation with Python}}.
\newblock \bibinfo{howpublished}{\url{https://github.com/wkentaro/labelme}}.
%Type = Article
\bibitem[{Wang et~al.(2020)Wang, Sun, Cheng, Jiang, Deng, Zhao, Liu, Mu, Tan, Wang et~al.}]{wang2020deep}
\bibinfo{author}{Wang, J.}, \bibinfo{author}{Sun, K.}, \bibinfo{author}{Cheng, T.}, \bibinfo{author}{Jiang, B.}, \bibinfo{author}{Deng, C.}, \bibinfo{author}{Zhao, Y.}, \bibinfo{author}{Liu, D.}, \bibinfo{author}{Mu, Y.}, \bibinfo{author}{Tan, M.}, \bibinfo{author}{Wang, X.}, et~al., \bibinfo{year}{2020}.
\newblock \bibinfo{title}{Deep high-resolution representation learning for visual recognition}.
\newblock \bibinfo{journal}{IEEE transactions on pattern analysis and machine intelligence} \bibinfo{volume}{43}, \bibinfo{pages}{3349--3364}.
%Type = Article
\bibitem[{Weimerskirch et~al.(2000)Weimerskirch, Barbraud and Lys}]{weimerskirch2000sex}
\bibinfo{author}{Weimerskirch, H.}, \bibinfo{author}{Barbraud, C.}, \bibinfo{author}{Lys, P.}, \bibinfo{year}{2000}.
\newblock \bibinfo{title}{Sex differences in parental investment and chick growth in wandering albatrosses: fitness consequences}.
\newblock \bibinfo{journal}{Ecology} \bibinfo{volume}{81}, \bibinfo{pages}{309--318}.
%Type = Misc
\bibitem[{{Xiashu Technology}(2024)}]{xiashu2024automatic}
\bibinfo{author}{{Xiashu Technology}}, \bibinfo{year}{2024}.
\newblock \bibinfo{title}{Automatic chicken sexing system}.
\newblock \bibinfo{howpublished}{\url{https://www.xiashutech.com/en/pd.jsp?fromColId=119&id=3\#_pp=119_1373}}.
\newblock \bibinfo{note}{Accessed: Sep. 2024}.
%Type = Inproceedings
\bibitem[{Zhang and Wang(2011)}]{zhang2011hierarchical}
\bibinfo{author}{Zhang, G.}, \bibinfo{author}{Wang, Y.}, \bibinfo{year}{2011}.
\newblock \bibinfo{title}{Hierarchical and discriminative bag of features for face profile and ear based gender classification}, in: \bibinfo{booktitle}{2011 International joint conference on biometrics (IJCB)}, \bibinfo{organization}{IEEE}. pp. \bibinfo{pages}{1--8}.
%Type = Inproceedings
\bibitem[{Zhang(2018)}]{zhang2018improved}
\bibinfo{author}{Zhang, Z.}, \bibinfo{year}{2018}.
\newblock \bibinfo{title}{Improved adam optimizer for deep neural networks}, in: \bibinfo{booktitle}{2018 IEEE/ACM 26th international symposium on quality of service (IWQoS)}, \bibinfo{organization}{Ieee}. pp. \bibinfo{pages}{1--2}.

\end{thebibliography}

\end{document}